\documentclass[a4paper,fleqn]{cas-sc}

\usepackage[authoryear,longnamesfirst]{natbib}

\usepackage{subfigure} 
\usepackage{ragged2e}
\usepackage{setspace}

\def\tsc#1{\csdef{#1}{\textsc{\lowercase{#1}}\xspace}}
\tsc{WGM}
\tsc{QE}

\begin{document}
\let\WriteBookmarks\relax
\def\floatpagepagefraction{1}
\def\textpagefraction{.001}


\shorttitle{Cross-Modality Attentive Feature Fusion }    

\shortauthors{FANG Qingyun,WANG Zhaokui}  

\title [mode = title]{Cross-Modality Attentive Feature Fusion for Object Detection in Multispectral Remote Sensing Imagery }  



%


\author[1]{FANG Qingyun}



\ead{fqy17@mails.tsinghua.edu.cn}



\affiliation[1]{organization={School of Aerospace Engineering, Tsinghua University},
            city={Beijing},
            postcode={100084}, 
            country={China}}

\author[1]{WANG Zhaokui}

\cormark[1]


\ead{wangzk@tsinghua.edu.cn}


\credit{}


\cortext[1]{Corresponding author}



\begin{abstract}
\doublespacing

Cross-modality fusing complementary information of multispectral remote sensing image pairs can improve the perception ability of detection algorithms, making them more robust and reliable for a wider range of applications, such as nighttime detection.  
Compared with prior methods, we think different features should be processed specifically, the modality-specific features should be retained and enhanced, while the modality-shared features should be cherry-picked from the RGB and thermal IR modalities. 
Following this idea, a novel and lightweight multispectral feature fusion approach with joint common-modality and differential-modality attentions are proposed, named Cross-Modality Attentive Feature Fusion (CMAFF).
Given the intermediate feature maps of RGB and IR images, our module parallel infers attention maps from two separate modalities, common- and differential-modality, then the attention maps are multiplied to the input feature map respectively for adaptive feature enhancement or selection.
Extensive experiments demonstrate that our proposed approach can achieve the state-of-the-art performance at a low computation cost.

\end{abstract}


\begin{highlights}
\doublespacing
\item We propose a simple yet effective CMAFF module that can fuse the complementary information of multispectral remote sensing images with joint common-modality and differential-modality attentions.
\item  We confirm the effectiveness of our cross-modality fusion attention module through extensive ablation studies.
\item We design a new two-stream object detection network YOLOFusion for multispectral remote sensing images and verify its performance.
\end{highlights}


\maketitle

\doublespacing
\section{Introduction}\label{intro}

Object detection is a canonical task in computer vision, as well as in remote sensing. 
Object detection in remote sensing imagery deals with detecting instances of visual objects of certain classes, most of which are man-made,  buildings, airplanes, ships, vehicles, to name a few. 
This technology has been widely used in many civilian and military fields, such as port and airport flow monitoring, traffic diversion, urban planning, lost ship search and rescue.

Traditional machine learning (ML) schemes based on the encoding of handcrafted features (e.g., textures, color histogram, or more complex HOG \cite{AAdalal2005histograms}, SIFT \cite{AAlowe2004distinctive}, Haar \cite{AAviola2001rapid},ACF \cite{AAdollar2014fast}, etc.) can only generate shallow to middle features with limited representativity.
Recently, with the rapid development of deep learning (DL), convolutional neural networks (CNNs) have became a new and powerful approach for feature extraction and greatly improved the performance of object detection. 
Current CNN-based object detection methods could be roughly divided into two streams: two-stage schemes and one-stage schemes. The two-stage detector, such as R-CNN \cite{AAgirshick2014rich}, Fast R-CNN \cite{AAgirshick2015fast}, Faster R-CNN \cite{AAren2015faster} and other detectors \cite{AAcai2018cascade,AApang2019libra,AAli2019scale}, divide the detection into localization and recognition stages, having one more region-proposal step than single-stage detectors.
On the contrary, single-stage detectors, such as RetinaNet \cite{AAlin2017focal}, SSD \cite{AAliu2016ssd}, EfficientDet \cite{AAtan2020efficientdet}, YOLO \cite{AAredmon2016you}, etc., take advantage of regression to achieve localization, along with recognition.

Recent advances in optical sensor technology and CNN-based algorithms have encouraged detection performance in several remote sensing scenarios \cite{AAvan2018you,AAding2019learning,AApham2020yolo,AAqingyun2020efficient,AAhan2021align}.
However, the objects may appear under varying conditions of illumination, weather, resolution, and occlusions.
In particular, it cannot work at night, which vastly restricts the application of detection for visible-band sensors.
It has since been established that adopting multispectral imaging technology can improve the performance under such difficult conditions\cite{yang2019air}.
In addition to the visible spectrum, one additional well-suited modality is the thermal infrared (IR) spectrum.
Fusing the complementary modalities of RGB and IR can further improve the perceptibility of algorithms in object detection.

There are considerable papers that aim at multispectral pedestrian detection  \cite{AAhwang2015multispectral,AAliu2016multispectral,AApark2018unified,AAli2019illumination,AAzhang2021guided}, while multispectral object detection in remote sensing imagery has just started \cite{AAdhanaraj2020vehicle,AAsharma2021yolors}.
Naturally, how to design an efficient cross-modality feature fusion mechanism is the most important step in the entire multispectral detection algorithm.
In prior literature, the fusion mechanism using simple operations(e.g., concatenation, element-wise addition, and element-wise cross product) cannot fully exploit the inherent complementary between different modalities.
Worse still, these coarse approaches may increase the difficulty of network learning, and aggravate the imbalance of the modalities, which will result in performance degradation.

\begin{figure}[htbp]
	\centering
		\includegraphics[width=0.7\linewidth]{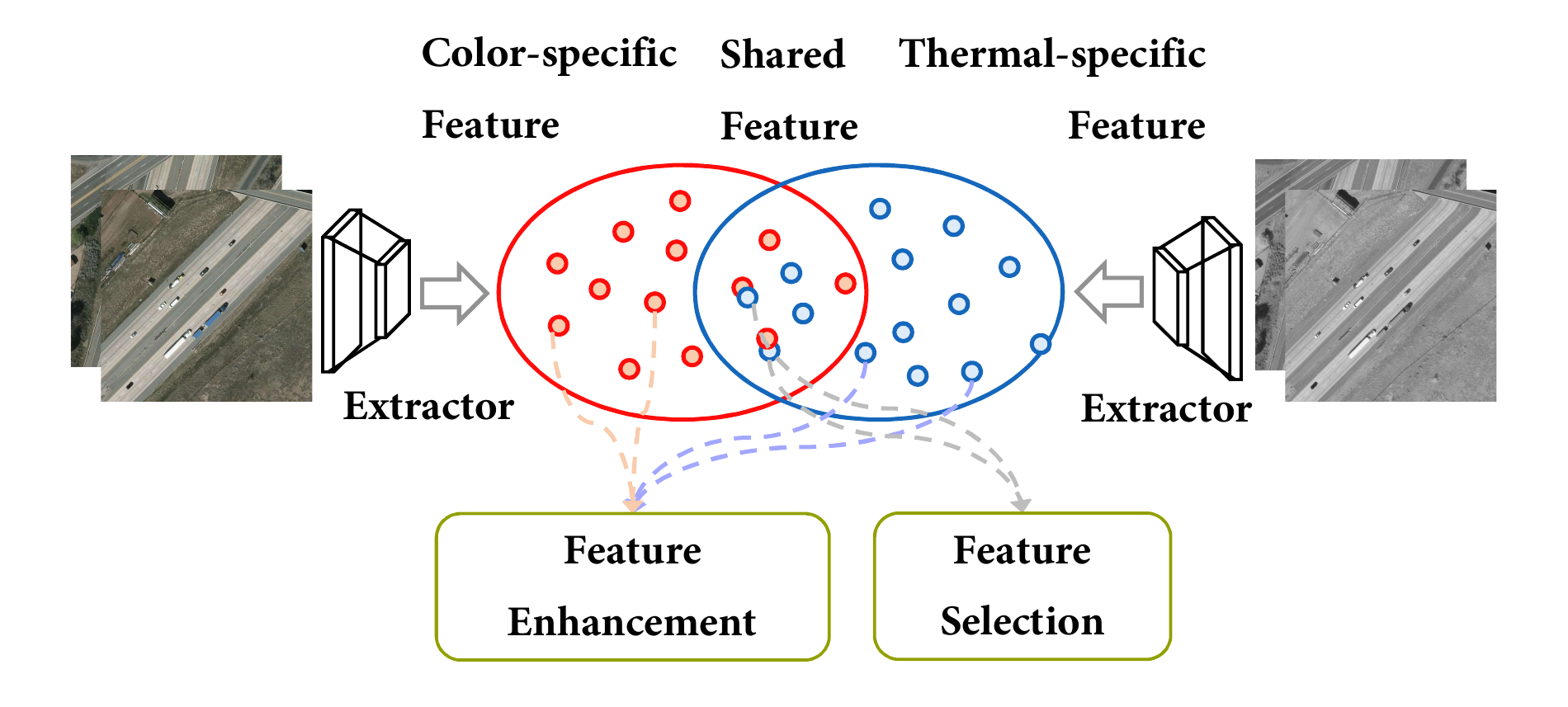}
	  \caption{Illustration of our idea for cross-spectral feature representation learning. Direct merge will inevitably introduce redundant features and a selection mechanism is designed to cherry-pick the better and filter the worse. Additionally, feature enhancement helps the network learn the modality discrepancy. }\label{fig_idea}
\end{figure}


Due to this difficulty, our solution is to decouple different features from source and process them.
The modality-specific features (e.g. color, thermal) should be retained and enhanced, while the modality-shared features (e.g. shape, spatial relationship) should be cherry-picked from the two modalities.
Figure \ref{fig_idea} illustrates our idea.
Two feature extractors are constructed to encode each spectral image into a common feature space.
The entire common feature space is divided into three sub-spaces, color-specific feature space, shared feature space, and thermal-specific feature space.
We adapt the feature enhancement for color- and thermal-specific features so that the network learns more about the different characteristics of the two modalities.
Meanwhile, feature selection is performed on the shared features to mitigate the effect of redundant features.
In this way, our algorithm can effectively utilize both shared and specific information for each sample.

For the above considerations, we propose a novel and lightweight multispectral fusion approach with joint common-modality and differential-modality attentions in this paper, named \textit{Cross-Modality Attentive Feature Fusion}  (CMAFF). 
To demonstrate that our proposed CMAFF is powerful, we combine it with the lightweight YOLOv5-small \cite{AAjocher2020ultralytics} detector called YOLOFusion.
YOLOFusion is a unified end-to-end multi-spectral detection method.
And to the best of our knowledge, this is the first algorithm that detects targets with attention fusion in the area of multispectral remote sensing imagery.

The rest of this paper is organized as follows. 
Some related work is introduced in section \ref{sec_related}.
The implementation details of our CMAFF module is given in section \ref{sec_method}.
We provide an extensive ablation study and visualization results to discuss the reasons of the performance improvements in section \ref{sec_exper}.
Finally, the conclusions are drawn in section \ref{sec_conclu}.

\section{Related work}\label{sec_related}

\begin{figure*}[htbp]
    \centering
    \subfigure[early fusion]{
    \includegraphics[height=4.5cm]{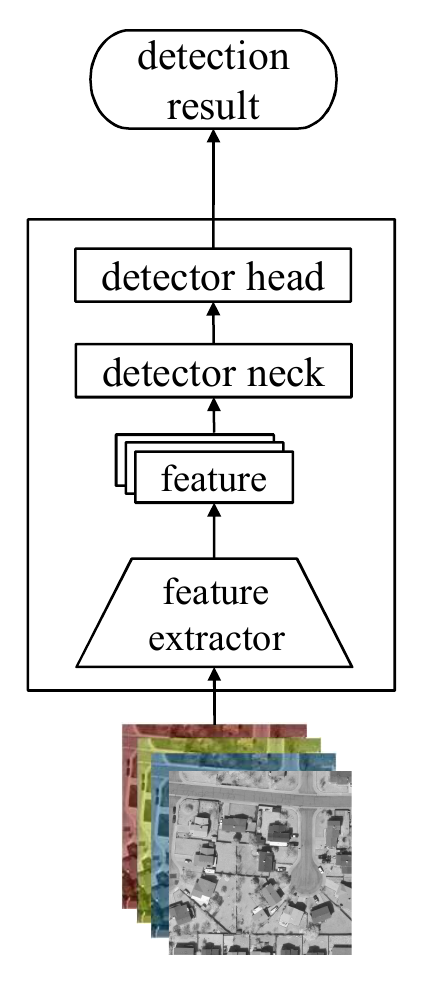}
    }
    \quad
    \subfigure[middle fusion]{
    \includegraphics[height=4.5cm]{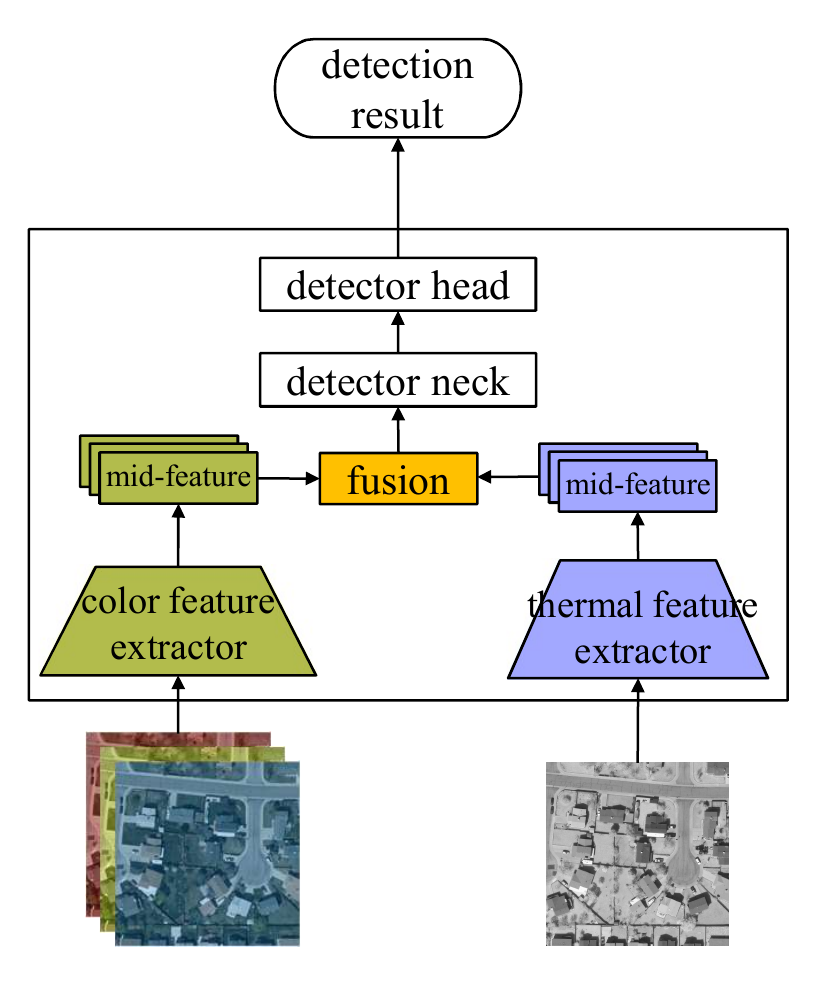}
    }
    \quad
    \subfigure[late fusion]{
    \includegraphics[height=4.5cm]{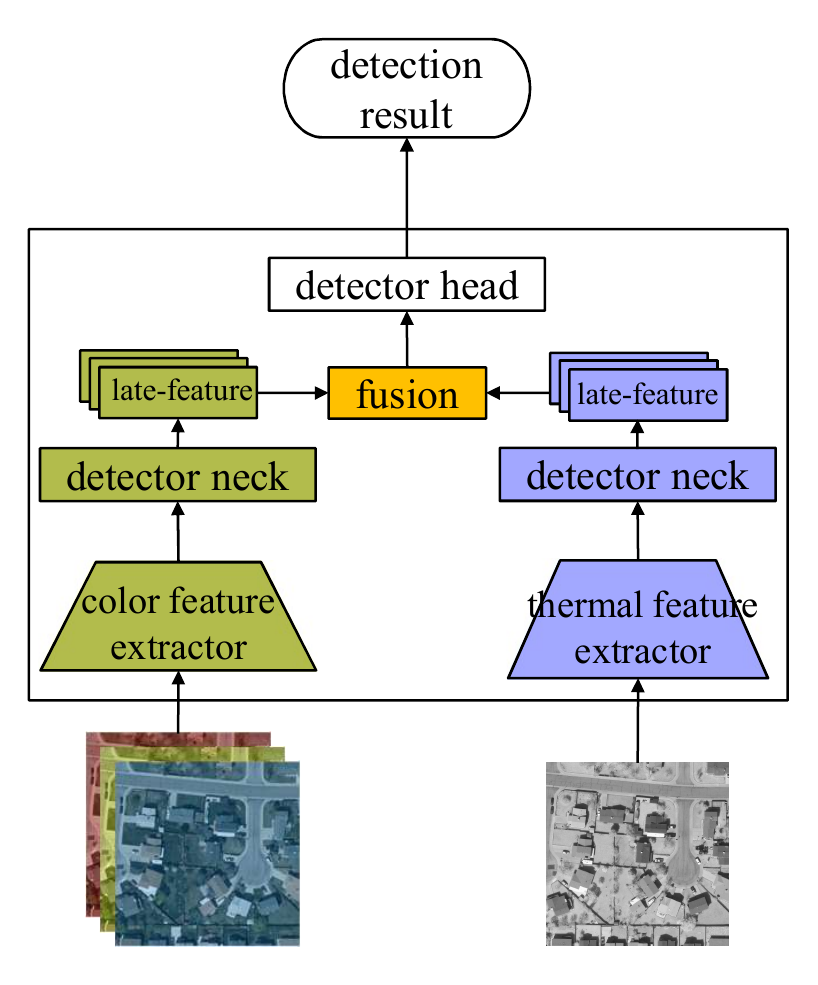}
    }
    \quad
    \subfigure[score fusion]{
    \includegraphics[height=4.5cm]{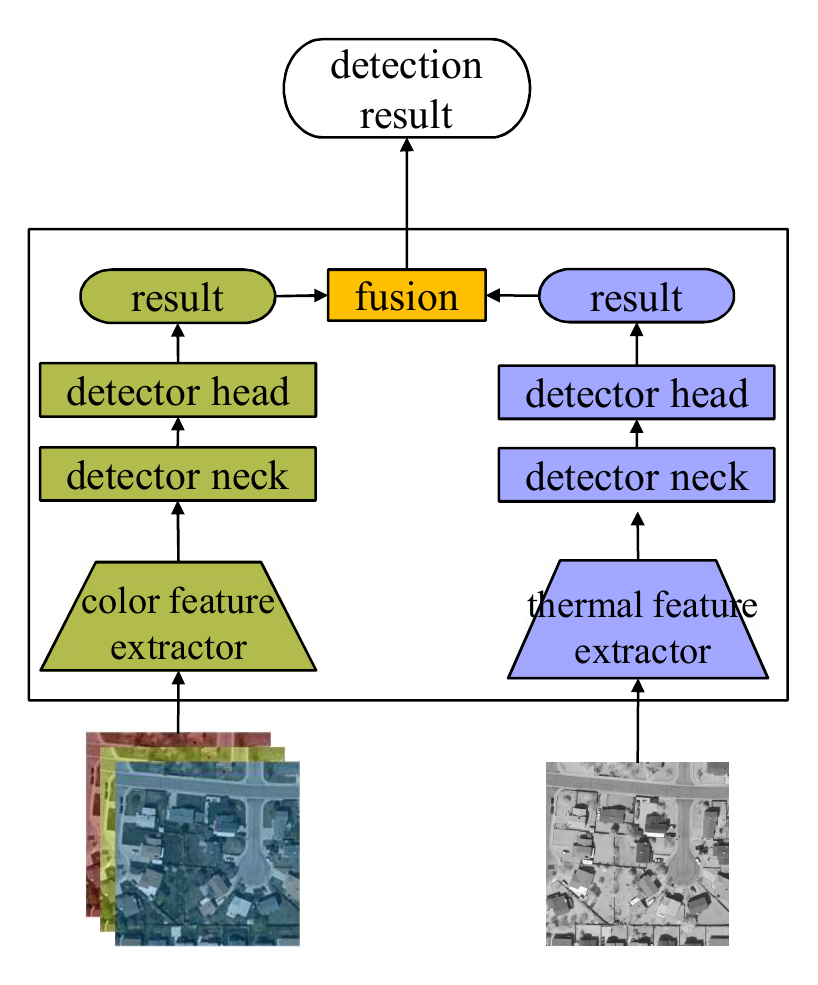}
    }

    \caption{
    Four fusion strategies.
    (a) Early fusion simply performs pixel-level addition operation on RGB and IR images, or concatenates them as a four-channel input to integrate color and thermal information, and then follows the training of a mono-modality (RGB-based) detector.
    (b) Middle fusion fuses the features computed from two independent feature extractors and inputs the fused features into the shared detector neck and detector head (see the detector neck and detector head in YOLOV4 \cite{AAbochkovskiy2020yolov4} and YOLOv5 \cite{AAjocher2020ultralytics} for details).
    (c) Late fusion fuses late features after detector necks.
    (d) Score fusion usually fuses the confidence scores and prediction bounding boxes of two mono-modality detectors in the final decision stage. (Best viewed in color.)
    }
    \label{fig_fusionform}
\end{figure*}

The related works on object detection in multispectral remote sensing imagery are presented in this section, including advanced detection algorithms and multispectral fusion approaches.

\subsection{Advanced detection algorithms}\label{subsec_advance}

Advanced CNN-based detectors train deep networks over publicly large-scale and high‐quality natural image datasets, such as Microsoft Common Objects in Context (MS COCO) \cite{AAlin2014microsoft}  and PASCAL Visual Object Classes (PASCAL VOC) \cite{AAeveringham2010pascal}. 
In view of the concise network architecture, super-fast running speed and outstanding performance of the YOLO algorithm, it has drawn much attention from academia and industry since its introduction in 2016. 
Until now, new variants of the YOLO series are still being proposed and developed (e.g., YOLOv2 \cite{AAredmon2017yolo9000}, YOLOv3 \cite{AAredmon2018yolov3}, YOLOv4 \cite{AAbochkovskiy2020yolov4}, YOLOv5 \cite{AAjocher2020ultralytics} and YOLOX \cite{AAge2021yolox}). 
As mentioned in Section \ref{intro}, the YOLOv5-small detector is adopted as a basic framework in this work for exploring multimodal detection. 
Meanwhile, choosing a lightweight one-stage algorithm is also facilitated to the future implementation of online real-time multispectral detection on satellites and UAVs.
Note that our explored approach is general and can apply to other mainstream detectors.

\subsection{Multispectral fusion approaches}\label{subsec_multispectral}
How to fuse the information of two modalities is the core problem in multispectral object detection. 
According to the different stages of fusion \cite{AAliu2016multispectral, AAchen2021multimodal}, multispectral fusion approaches can be categorized into four forms: early fusion, middle fusion, late fusion and score fusion, as shown in Fig.~\ref{fig_fusionform}. 
Several explored work \cite{AAliu2016multispectral, AAcao2021attention} has shown that middle fusion is overwhelmingly outperforming than other three fusion forms, regardless of whether the basic detector framework is single-stage or two-stage.
Therefore, in this work, the middle fusion strategy is adopted to detect interest targets in multispectral remote sensing imagery.

Besides the fusion strategy at different stages, the fine design of the fusion module is also crucial for inherent complementary learning of color and thermal modalities.
CIAN \cite{AAzhang2019cross} makes the recalibrated middle-level feature maps of two streams converge to a unified one under the guidance of cross-modality interactive attention and introduces the context enhancement blocks (CEBs) to further enhance contextual information. 
Two variations of novel Gated Fusion Units (GFU) are proposed in GFD-SSD \cite{AAzheng2019gfd}  to learn the combination of feature maps generated by the two SSD middle layers.
\cite{AAzhang2021guided} propose a fully adaptive multispectral feature fusion approach, which combines the intra-modality and inter-modality attention modules, allows the network to learn the dynamical weighting and fusion of multispectral features.
The MCFF module \cite{AAcao2021attention} based on channel-wise attention integrates the features from the color and thermal streams according to the illumination conditions to achieves state-of-the-art performance.

Regardless of the design, a guiding idea of the fusion module is to retain and enhance the reliable features in the two-modalities feature space, while suppressing the information delivered by redundant and degraded features.

\section{Method}\label{sec_method}
In this section, our one-stage multispectral detection approach YOLOFusion is introduced, including the overall detection architecture and the \textit{Cross-Modality Attentive Feature Fusion} (CMAFF) module.

\subsection{Overall YOLOFusion architecture}\label{subsec_yolofusion}
YOLOv5 \cite{AAjocher2020ultralytics} is a one-stage anchor-base detector that has been widely utilized in various industries due to its real-time inference speed.
Considering the future deployment of our multispectral object detection algorithm on edge devices, the small version of YOLOv5 (i.e., YOLOv5-small) is adopted as our baseline.

\begin{figure}[!htbp]
	\centering
		\includegraphics[width=1.0\linewidth]{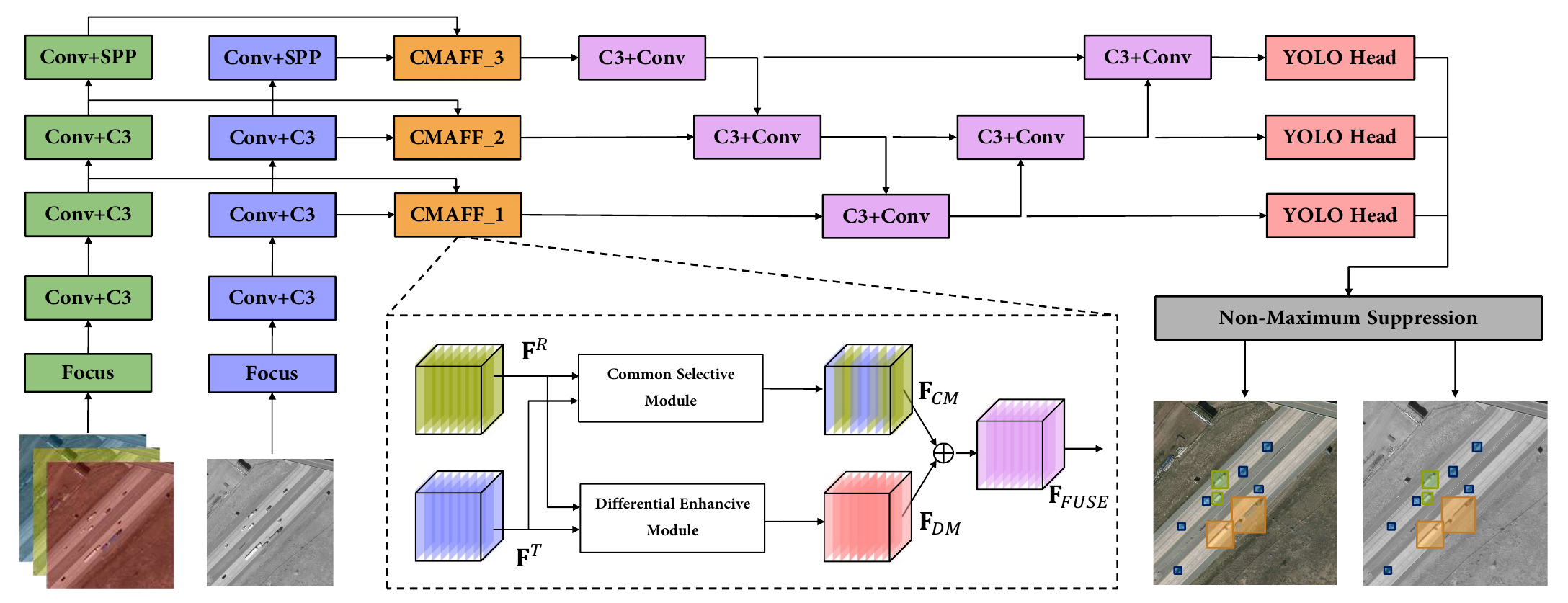}
	  \caption{YOLOFusion architecture.
 Our YOLOFusion consists of 5 parts in total. The first is a dual-stream feature extraction network, which extracts RGB modal features and IR modal features. Then the CMAFF module fuses the two-modality extracted features in a divide-and-conquer manner.The PANet-like feature pyramid aggregates features of different scales and depths, and further refines the spatial and semantic information of the features. Finally, the detection is completed by the detection head and the post-processing (NMS). }
	  \label{fig_yolofusion}
\end{figure}

In this work, we combine the YOLOv5-small with our fine-designed CMAFF module, called YOLOFusion.
Our YOLOFusion is a unified end-to-end multi-spectral detection network, and the details of its architecture are illustrated in Fig.~\ref{fig_yolofusion}. 
Compared with the original version, our algorithm constructs a two-stream convolutional neural network backbone, represented by the blue and green modules on the left in Fig.~\ref{fig_yolofusion}, to extract the feature maps of the RGB images and the IR images respectively.
And then, our proposed CMAFF module is taking place in the fusion stage with the ability to merge mid-level multispectral feature maps of color and thermal modality branches.
More specifically, fusion based on the CMAFF modules occurs in three different stage, allowing the fused features to contain multi-scale semantic information, as well as fine spatial details.
Furthermore,the proposed CMAFF module is plug-and-play and can be embedded in any type of two-stream CNN-based detectors as a fusion module.

\subsection{Cross-Modality Attentive Feature Fusion}\label{subsec_cmaff}
It is inadvisable to capture the cross-modality complementary representation by previous direct concatenation or summation scheme.
Both RGB and IR modalities have their own inherent characteristics which are mixed with cues and noises.
While simple fusion schemes such as linear combination or concatenation are lacking in clarity to mine and comb the modality-shared and modality-specific features.

Following the idea of divide and conquer, and inspired by the differential amplifier in which the differential-mode signals are amplified and the common-mode signals are suppressed \cite{AAzhou2020improving}, our CMAFF module consists of two parts,
differential enhancive module and common selective module, as shown in Fig.~\ref{fig_yolofusion}.
According to the principle of the differential amplifier, the RGB convolution feature maps $\mathbf{F}^{R}$ and the IR convolution feature maps $\mathbf{F}^{T}$ can be represented with common-modality component and differential-modality component at each channel as follows:
\begin{align}
\label{eq_fr} &\mathbf{F}^{R}=\frac{\mathbf{F}^{R}+\mathbf{F}^{R}}{2}+\frac{\mathbf{F}^{T}-\mathbf{F}^{T}}{2}=\frac{\mathbf{F}^{R}+\mathbf{F}^{T}}{2}+\frac{\mathbf{F}^{R}-\mathbf{F}^{T}}{2} = \frac{\mathbf{F}^C}{2} + \frac{\mathbf{F}^D}{2} \\
\label{eq_ft}& \mathbf{F}^{T}=\frac{\mathbf{F}^{T}+\mathbf{F}^{T}}{2}+\frac{\mathbf{F}^{R}-\mathbf{F}^{R}}{2}=\frac{\mathbf{F}^{R}+\mathbf{F}^{T}}{2}+\frac{\mathbf{F}^{T}-\mathbf{F}^{R}}{2} = \frac{\mathbf{F}^C}{2} - \frac{\mathbf{F}^D}{2} \\
& \mathbf{F}^C = \mathbf{F}^{R}+\mathbf{F}^{T} \\
& \mathbf{F}^D = \mathbf{F}^{R}-\mathbf{F}^{T} 
\end{align}
where the $\mathbf{F}^C$ reflects the common features and the $\mathbf{F}^D$ reflects the distinct features captured by two modalities.

Intuitively, when learning cross-modal complementary, the network needs to learn not only the differences between RGB and IR modalities but also their commonalities and consistency.
Consequently, our CMAFF module does not suppress the common-modality feature.
On the contrary, the most effective channel features of the RGB and IR modalities are selected by the common-modality feature and remixed into a new feature.
In this way, we believe that based on the common-modality selective module, the network can extract modality-shared features, while  modality-specific features can be obtained with the differential-modality enhancive module.

After the refinement with differential enhancive and common selective  modules, two more informative and robust feature maps are generated and added to complete the final fusion.

\begin{figure*}[!htbp]
	\centering
		\includegraphics[width=1.0\linewidth]{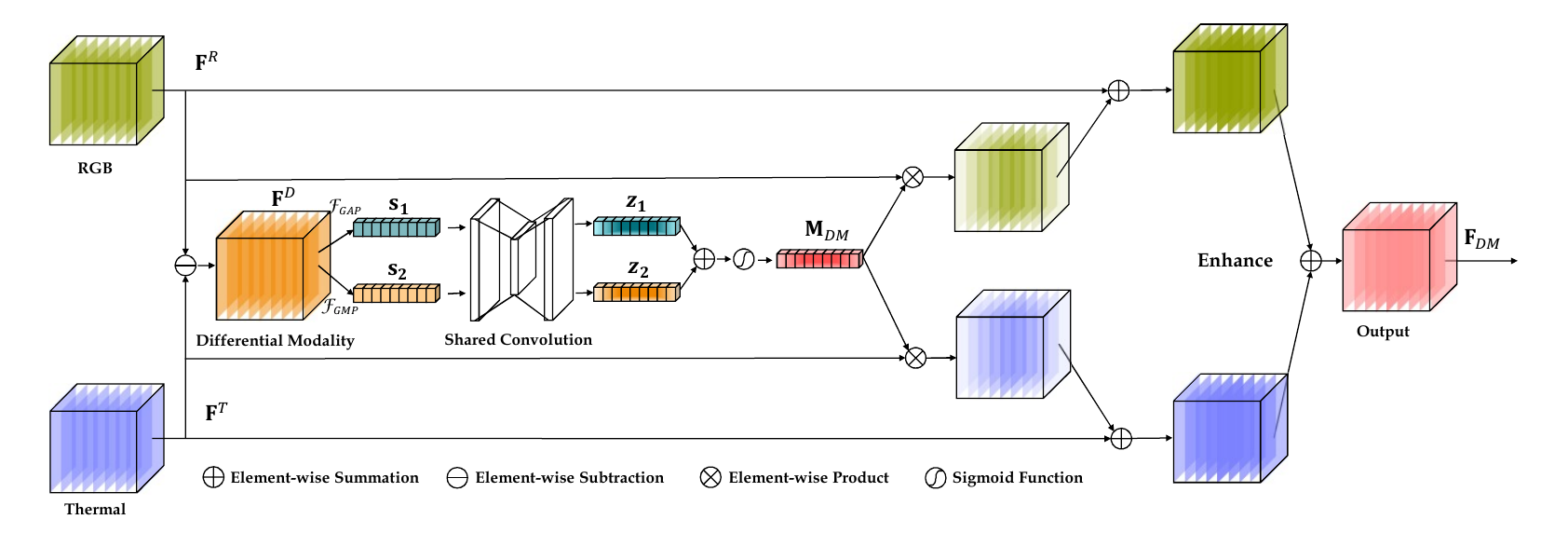}
	  \caption{Differential Enhancive Module.
	  Based on the differential-modality features, this module use both average-pooled and max-pooled features simultaneously to obtain the global receptive field, improve the spatial information aggregating ability, and ultimately greatly enhance representation power of networks.
	  }\label{fig_DMEA}
\end{figure*}

\textbf{Differential Enhancive} module is inspired by the differential mode signals being amplified in differential amplifier circuits.  
This module aims at enhancing the thermal or visible features by differential-modality features.
As illustrated in Fig.~\ref{fig_DMEA}, the main idea behind our module is leveraging the channel-wise attention weighting mechanism to enhance features from differential modality.
Specifically, given the intermediate RGB convolution feature maps $\mathbf{F}^R \in \mathbb{R}^{C \times H \times W}$ and thermal convolution feature maps $\mathbf{F}^T  \in \mathbb{R}^{C \times H \times W} $, we obtain the differential feature maps $\mathbf{F}^D  \in \mathbb{R}^{C \times H \times W} $ by direct subtraction of two modalities first. 
Second, our differential enhancive module infers the attention maps of channel dimensions  $\mathbf{M}_{DM} \in \mathbb{R}^{C \times 1 \times 1}$ based on differential-modality feature maps $\mathbf{F}_D$, then the attention maps are multiplied to the each input feature maps for adaptive feature refinement. 
Thirdly, the refined feature maps are added by the input feature maps to get the enhanced features maps of each modality.
Finally, Sum the two enhanced features maps to get the output of the differential enhancive module $\mathbf{F}_{DM} \in \mathbb{R}^{C \times H \times W}$.

This procedure can be formalized as:
\begin{equation}
    \mathbf{F}_{DM} = \mathbf{F}^R \otimes (1+ \mathbf{M}_{DM}) + \mathbf{F}^T \otimes (1+ \mathbf{M}_{DM})
\label{eq_f_dm}
\end{equation}
 where,
\begin{align}
    & \mathbf{M}_{DM} = \sigma( \mathbf{z_1} + \mathbf{z_2} ) \\
    & \mathbf{z_1} = \mathcal{F}_{SC}(\mathbf{s_1} ) \\
    & \mathbf{z_2} = \mathcal{F}_{SC}(\mathbf{s_2} ) \\
    & \mathbf{s_1} = \mathcal{F}_{GAP}(\mathbf{F}^D)=\frac{1}{H \times W} \sum_{i=1}^{H} \sum_{j=1}^{W} \mathbf{F}^D(i, j) \\
    & \mathbf{s_2} = \mathcal{F}_{GMP}(\mathbf{F}^D)= \operatorname{max}(\mathbf{F}^D(i, j))
\end{align}

Superscripts $R$ and $T$ denote the RGB (R) and Thermal (T) modality; 
superscripts $D$ denote Differential(D) modality;
subscript $DM$ indicates that the variables belong to the differential enhancive module, which is used to distinguish the variables in the common selective module later.
$\otimes$ denotes the element-wise multiplication; 
$\sigma$ represents the sigmoid function; 
$\mathcal{F}_{SC}$ is considered as the shared convolution operation; 
$\mathcal{F}_{GAP}$ and  $\mathcal{F}_{GMP}$ refer to Global Average Pooling and Global Max Pooling, respectively.
$\mathbf{z_1} \in \mathbb{R}^{C \times 1 \times 1} $, $\mathbf{z_2} \in \mathbb{R}^{C \times 1 \times 1} $, $\mathbf{s_1} \in \mathbb{R}^{C \times 1 \times 1}$ and $\mathbf{s_2} \in \mathbb{R}^{C \times 1 \times 1}$ are four intermediate variables.

It is noteworthy that the refined feature maps are added as a residual that formulates the modality-specific feature enhancement learning as residual learning.
With residual mapping, the differential enhancive module can avoid directly impacting the thermal or RGB streams.

\begin{figure*}[!htbp]
	\centering
		\includegraphics[width=1.0\linewidth]{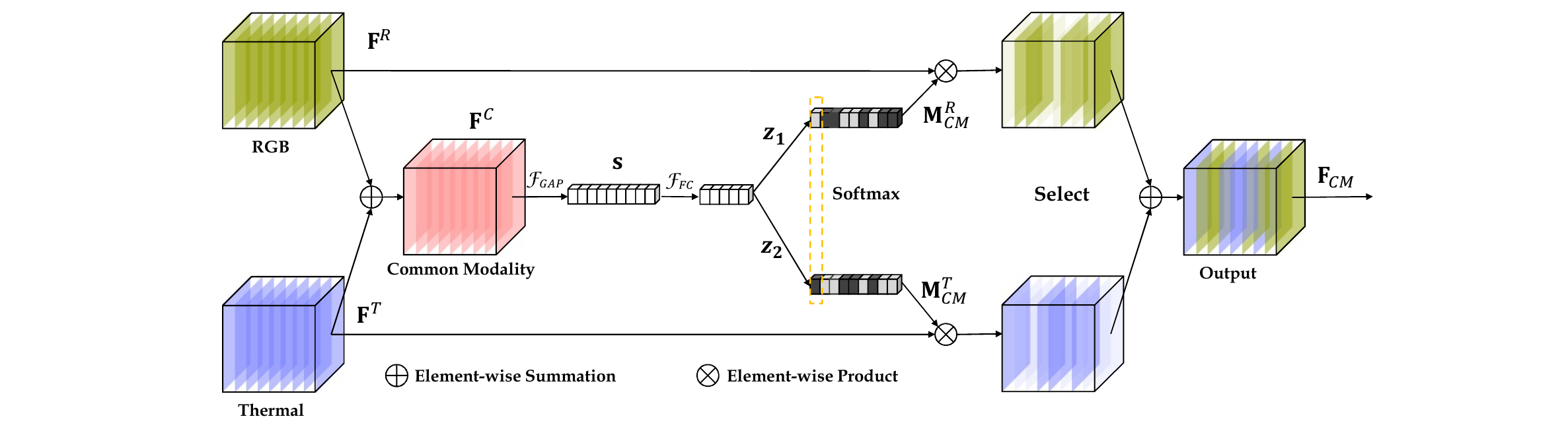}
	  \caption{Common Selective Module. 
	  Based on the common-modality features, this module utilizes softmax function to achieve a channel-wise selection mechanism, thereby picking effective features while filtering redundant features.
	  }\label{fig_CMSA}
\end{figure*}

\textbf{Common Selective} module is introduced to adaptively select two-modality channel features according to the common-modality feature and remixed into a new feature, as mentioned before. 
As illustrated in Fig.~\ref{fig_CMSA}, our common selective module can dynamically pick features with softmax attention which is guided by the information in RGB and thermal branches.
Similarly, given the RGB convolution feature maps $\mathbf{F}^R \in \mathbb{R}^{C \times H \times W}$ and thermal convolution feature maps $\mathbf{F}^T  \in \mathbb{R}^{C \times H \times W} $, the common-modality feature maps $\mathbf{F}^C  \in \mathbb{R}^{C \times H \times W} $ are extracted by direct summation first.
Then the common selective module compute the RGB feature attention maps  $\mathbf{M}_{CM}^R \in \mathbb{R}^{C \times 1 \times 1}$ and the thermal features attention maps $\mathbf{M}_{CM}^T \in \mathbb{R}^{C \times 1 \times 1}$.
Next, multiply the attention maps of RGB features and the attention maps of thermal features with their input respectively. 
Finally, the output of the common selective module $\mathbf{F}_{CM} \in \mathbb{R}^{C \times H \times W}$ is obtained by adding the output results of the previous step.

This procedure can be formulated as:
\begin{equation}
    \mathbf{F}_{CM} = \mathbf{F}^R \otimes \mathbf{M}_{CM}^R + \mathbf{F}^T \otimes \mathbf{M}_{CM}^T 
\label{eq_f_cm}
\end{equation}
where
\begin{align}
 & \mathbf{M}_{CM}^R, \mathbf{M}_{CM}^T = \mathcal{F}_{softmax}(\mathbf{z_1},\mathbf{z_2}) \\
 & \mathbf{z_1} = \mathcal{F}_{FC_1}( \mathbf{s} ) \\
 & \mathbf{z_2} = \mathcal{F}_{FC_2}(\mathbf{s} ) \\
 & \mathbf{s} = \mathcal{F}_{GAP}(\mathbf{F}^{C})=\frac{1}{H \times W} \sum_{i=1}^{H} \sum_{j=1}^{W} \mathbf{F}^C(i, j)
\end{align}

Here, superscript $C$ denotes the Common (C) modality;
subscripts $CM$ indicates that the variables belong to the common selective module, which is similar with previous subscripts $DM$;
$\mathcal{F}_{softmax}$ represents the softmax operation, so the corresponding elements of the vector $\mathbf{M}_{CM}^R$ and the vector $\mathbf{M}_{CM}^T$ sum to 1;
$\mathcal{F}_{FC_1} $ and $\mathcal{F}_{FC_2} $ refer to two two-layer fully connected (FC) networks;
$\mathbf{z_1} \in \mathbb{R}^{C \times 1 \times 1} $, $\mathbf{z_2} \in \mathbb{R}^{C \times 1 \times 1} $, $\mathbf{s} \in \mathbb{R}^{C \times 1 \times 1}$ are three intermediate variables.
It is worth noting that $\mathcal{F}_{FC_1} $ and  $\mathcal{F}_{FC_2} $ share the weight of the first layer, that is, the output of the first layer of these two FC networks is the same.

Moreover, the parameter amount of a vanilla FC network is enormous and its cost is unacceptable.
Therefore, for better efficiency, the two FC networks share the weight of the first layer, while the shared layer reduces the dimensionality to 1/32 of the input vector $\mathbf{s}$.

In our view, the sparkle of our common selective module is that it takes advantage of the normalization of the softmax function to reassign weights to the feature channels, thereby screening the RGB and thermal modality features. 
With this mechanism of adaptively selecting channel features from two modalities, the network can avoid introducing numerous redundant features, thus focusing on the most reliable modality-shared features.

\textbf{Combining differential enhancive module and common selective module} is the final step to complete fusion of RGB and thermal modalities. 

It can be formulated as
\begin{equation}
\begin{split}
        \mathbf{F}_{FUSE} & = \mathbf{F}_{DM} + \mathbf{F}_{CM} \\
         & = \mathbf{F}^R \otimes (1+ \mathbf{M}_{DM}) + \mathbf{F}^T \otimes (1+ \mathbf{M}_{DM}) + \mathbf{F}^R \otimes \mathbf{M}_{CM}^R + \mathbf{F}^T \otimes \mathbf{M}_{CM}^T  \\
         & = \mathbf{F}^R \otimes (1+ \mathbf{M}_{DM} +  \mathbf{M}_{CM}^R ) + \mathbf{F}^T \otimes (1+ \mathbf{M}_{DM} + \mathbf{M}_{CM}^T) 
\end{split}
\label{eq_fuse}
\end{equation}

As illustrated in the final simplified expression of Eq.~\eqref{eq_fuse},  each modality feature ($\mathbf{F}^R $ or $\mathbf{F}^T$) has undergone the joint refinement of two attention maps ($\mathbf{M}_{DM}$ and $\mathbf{M}_{CM}$). 
On the one hand, $\mathbf{M}_{CM}$ uses the common modality to select features. 
On the other hand, $\mathbf{M}_{DM}$ uses differential modality to enhance features.
In addition, the 1 in the formula means that our CMAFF module retains the original features of the two modalities.
The two attention maps are added as residual branches, which avoid the information loss caused by destroying the original feature.

\begin{figure}[htbp]
    \centering
    \subfigure[Parallel Manner]{
    \includegraphics[height=4cm]{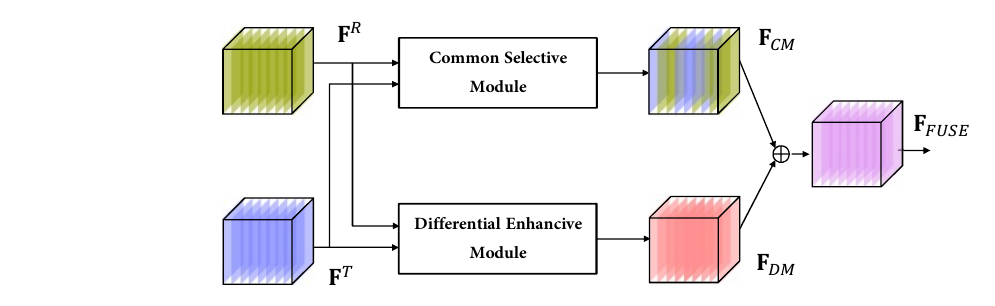}
    }
    \subfigure[Common-first Sequential Manner ]{
    \includegraphics[height=4cm]{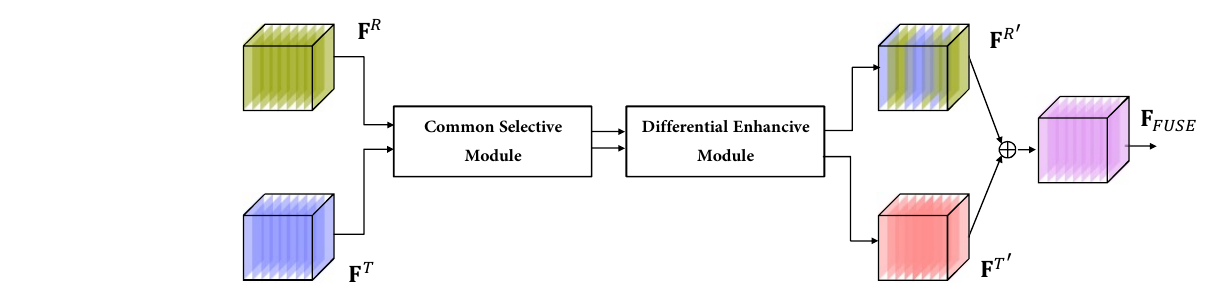}
    }
    \subfigure[Differential-first Sequential Manner]{
    \includegraphics[height=4cm]{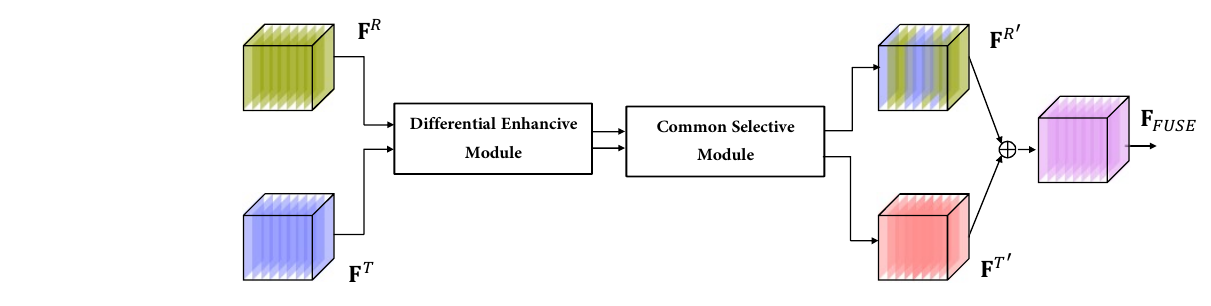}
    }

    \caption{
    The three arrangements of attention modules include the parallel manner, the common-first sequential manner, and the differential-first sequential manner. Note that in the sequential arrangement, both the differential enhancive module and the common selective module become dual-input dual-output. 
    }
    \label{fig_arrangement}
\end{figure}

\textbf{Arrangement of attention modules}. 
According to formulas \eqref{eq_f_dm}, \eqref{eq_f_cm} and \eqref{eq_fuse}, we can move the summation step of the differential enhancive module and the common selective module to the final combining step, then both two modules become dual-input and dual-output modules.
At this time, the differential enhancive module and the common selective module can be placed in a parallel or sequential manner, as illustrated in Fig.~\ref{fig_arrangement}.
Base on our work, the parallel arrangement gives a better result than a sequential arrangement. 
As for the arrangement of the sequential process, there are common-first and differential-first two orders.
We will discuss experimental results on arrangement of attention modules in section~\ref{subsubsec_CMAFF_architectures}.

\section{Experiment}\label{sec_exper}
In this section, we conduct experiments on the Vehicle Detection in Aerial Imagery (VEDAI) dataset \cite{AArazakarivony2016vehicle} to evaluate the performance of the proposed method.
Moreover, we attempt to interpret the reasons for improvements by visualizing the predicted attention masks. 
Finally, we provide speed and parameters analysis.

\subsection{Dataset}

The VEDAI dataset serves as a public database for evaluating the detection of small vehicles in multispectral remote images under various backgrounds such as woods, cities, roads, parking lot, construction sites or fields. 
Each image is available in RGB and IR modalities and the images in both modalities capture the same scene and are aligned with each other.
The dataset includes nine vehicle categories (car, truck, pickup, tractor, camper, ship, van, plane, other) for a total of more than 3700 annotated targets in more than 1200 images in each resolution, 1024 × 1024 or 512 × 512.
There is an average of 5.5 vehicles per image, and they occupy about 0.7\% of the total pixels of the images.
The spatial resolutions are 12.5 cm for the 1024 × 1024 images and 25 cm for the 512 × 512 images.
What’s more, 1024 × 1024 images are selected for training and test in our experiments.
As with the verification method of the VEDAI dataset provider\cite{AArazakarivony2016vehicle}, we adopt the 10-fold cross-validation protocol.

The annotations in the original dataset is in the format of a four-corner coordinate rotating box ($x_1, y_1; x_2, y_2; x_3, y_3; x_4, y_4$).
We convert these annotations to the horizontal-box format($x_{c}$,$y_c$,$w$,$h$) as follows:
\begin{itemize}
    \item the $x$ coordinate of the normalized center point, $x_c = \frac{\max(x1, x2, x3, x4) + \min(x1, x2, x3, x4)}{2\times1024}$;
    \item the $y$ coordinate of the normalized center point, $y_{c} = \frac{\max(y1, y2, y3, y4) + \min(y1, y2, y3, y4)}{2\times1024}$;
    \item the normalized width, $w = \frac{\max(x1, x2, x3, x4) - \min(x1, x2, x3, x4)}{1024}$;
    \item the normalized width, $h = \frac{\max(y1, y2, y3, y4) - \min(y1, y2, y3, y4)}{1024}$.
\end{itemize}

\subsection{Evaluation metrics}
In order to measure the performance of our CMAFF module and compare it to other methods, Precision, Recall, $\text{AP}_{0.5}$ (Average Precision) and $\text{mAP}_{0.5}$ (mean Average Precision) indicators are adopted in this paper. 

Precision, Recall, AP and mAP can be formalized as:
\begin{align}
& \text {Precision}=\frac{TP}{TP+FP} \\
&\text {Recall}=\frac{TP}{TP+FN} \\
\label{eq_ap} &\text{AP}=\int_{0}^{1}\text { Precision } \text{d} (\text{Recall}) = \int_{0}^{1} \frac{TP}{TP+FP} \text{d} \frac{TP}{TP+FN} = \int_{0}^{1} P(r) \text{ d} r \\
\label{eq_mAP} &\text {mAP}=\frac{1}{n}\sum_{i=0}^{n} AP_i = \frac{1}{n}\int_{0}^{1} P_i(r) ~\mathrm{ d} r
\end{align}

Where $TP$ is true positive, which means that a predicted box by detectors and the ground truth (GT) meet the intersection over union (IoU) threshold;
otherwise, it will be considered as a false positive ($FP$). False negative ($FN$) means there is a true target, but the detector doesn't find it. 
Equation~\eqref{eq_ap} indicates that $\text{AP}$ is the integral of the Precision-Recall Curve (PRC) for each category. 
And $\text{AP}_{0.5}$ means AP at IoU=0.50.
$\text { mAP }_{0.5}$ computes the mean of all the AP values for all categories at IoU=0.50 in the Eq.~\eqref{eq_mAP}.
$\text{mAP}_{0.5:0.95}$ is the primary challenge metric, which means $\text {mAP}_{0.5}$ at IoU=0.50:0.05:0.95. It is much more strict than $\text { mAP}_{0.5}$.

\subsection{Implementation details}
\begin{figure}[htbp]
    \centering

    \includegraphics[height=4.5cm]{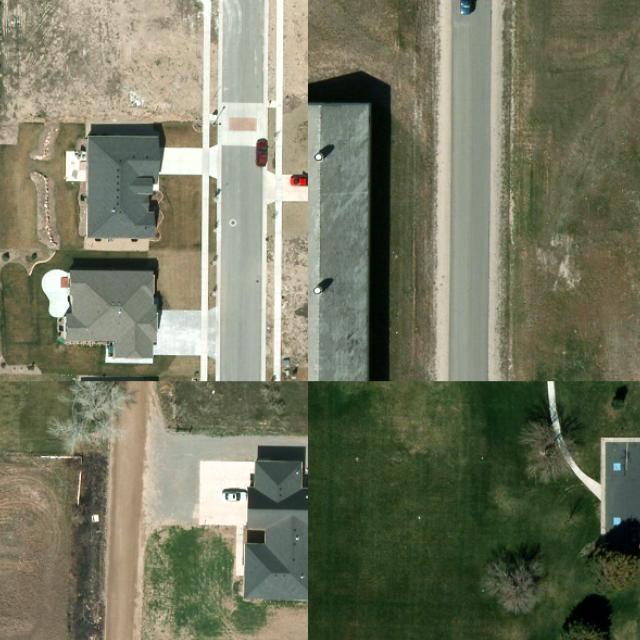}
    \includegraphics[height=4.5cm]{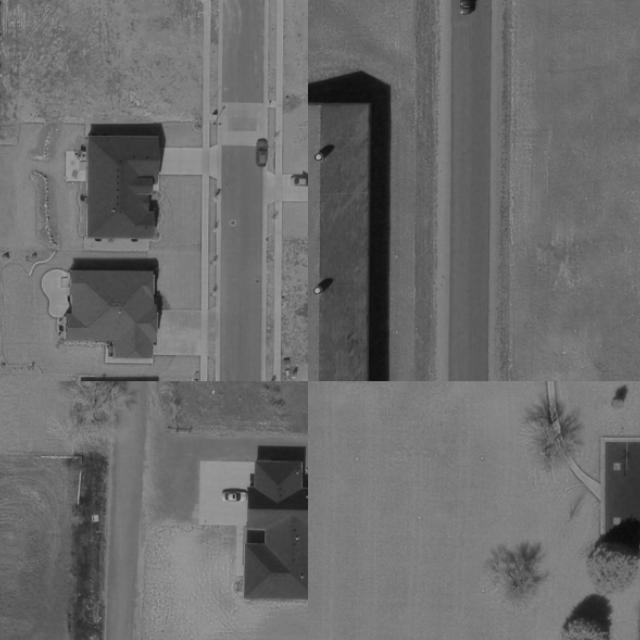}

    \vspace{0.08cm}

    \includegraphics[height=4.5cm]{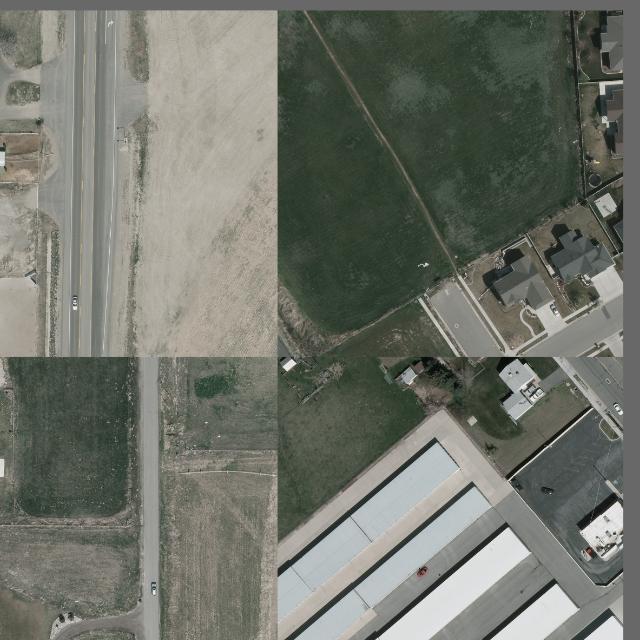}
    \includegraphics[height=4.5cm]{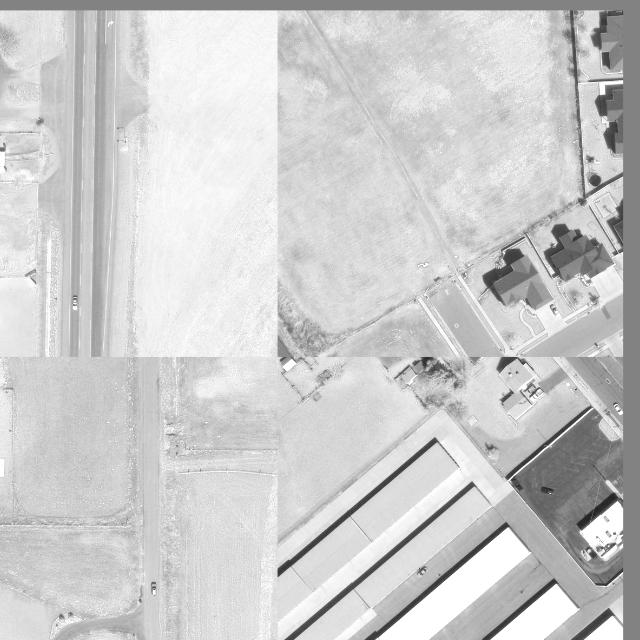}
    
    \vspace{0.08cm}
    
    \includegraphics[height=4.5cm]{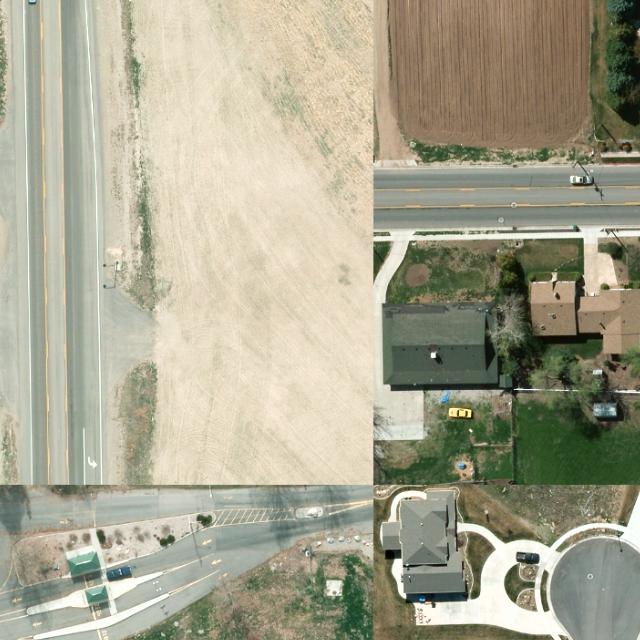}
    \includegraphics[height=4.5cm]{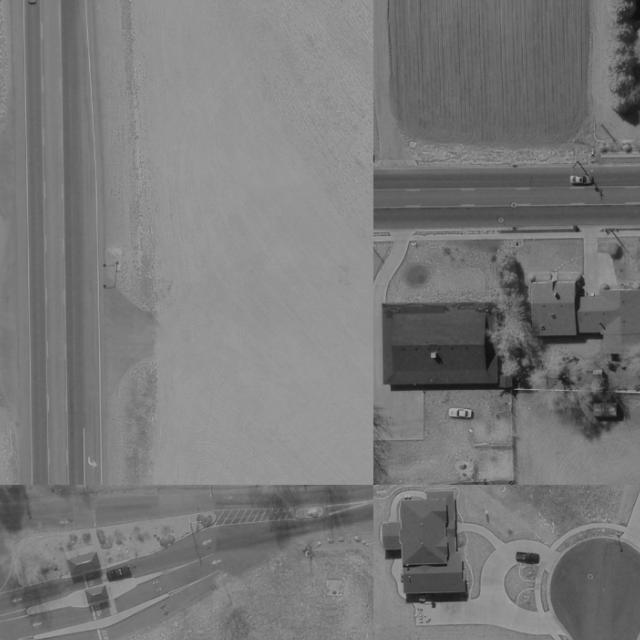}
    
    \vspace{0.08cm}
        
    \includegraphics[height=4.5cm]{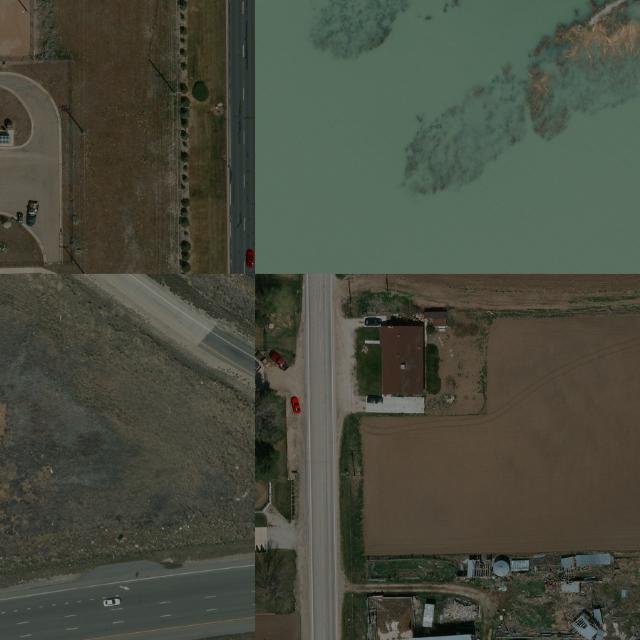}
    \includegraphics[height=4.5cm]{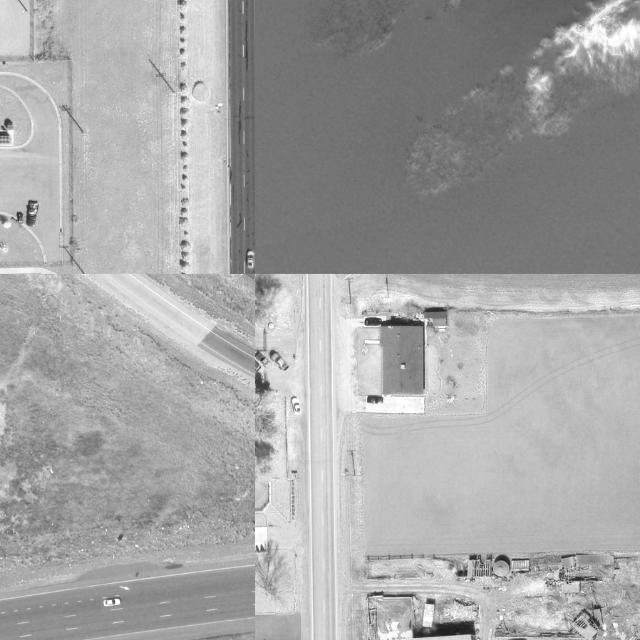}

    \caption{
    Four examples of the Mosaic data augmentation.
    }
    \label{fig_mosaic}
\end{figure}

\begin{figure}[!htbp]
	\centering
		\includegraphics[width=0.7\linewidth]{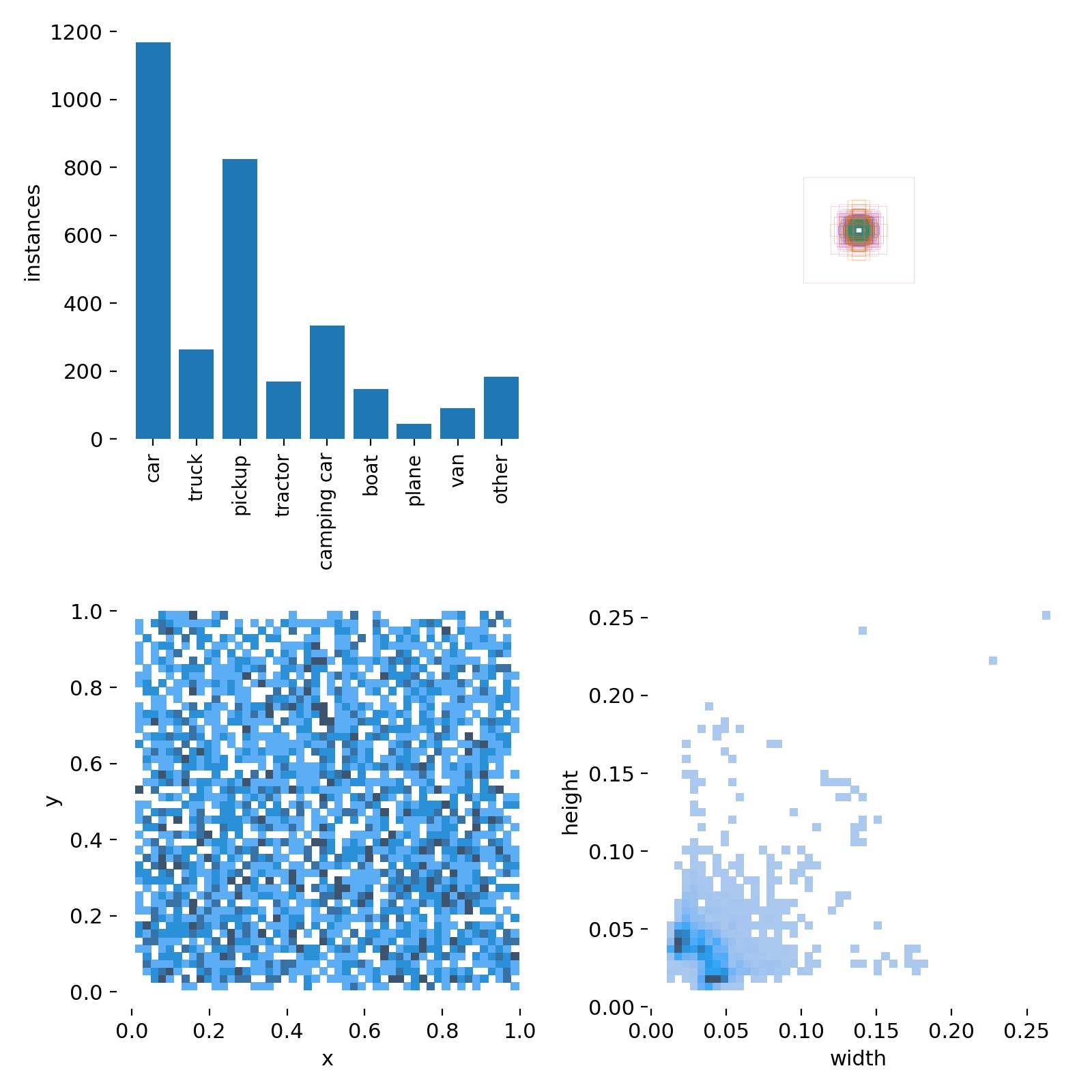}
	  \caption{Related statistics of the VEDAI dataset including the number of object instances of each class (upper left subgraph), the coordinate position distribution (bottom left subgraph) and the size distribution (bottom right subgraph) of the instances, and the visualization of the Ground Truth boxes (bottom right subgraph).}
	  \label{fig_label}
\end{figure}

All experiments with PyTorch \cite{AApaszke2019pytorch} on a single GPU (NVIDIA$^\text{®}$ TITAN RTX™). 
Our YOLOFusion detector uses YOLOv5-small as the base detector, which is pretrained on MS COCO \cite{AAlin2014microsoft}. 
The SGD optimizer with the initial learning rate of 1e-2 and the momentum of 0.937 is chosen in our experiments. 
The total epoch is 400, and the batch size is 32.
In both the training and test process, the input size images are 640×640.
In addition, we introduce the warmup trick \cite{AAzhang2019bag} to avoid violent oscillation at the beginning of training, so that the model can gradually stabilize under the low warmup learning rate. 
For data augmentation, we use the Mosaic approach as YOLOv4 \cite{AAbochkovskiy2020yolov4}, which mixes four training images. 
It allows the detection of objects outside their normal context. 
Note that when using data augmentation, the RGB and IR images should be processed the same, otherwise the input images (a pair of RGB and IR image) will be misaligned, causing the model to fail to converge. 
Figure \ref{fig_mosaic} illustrates four pairs of RGB and IR images augmented by the Mosaic approach.

As illustrated in Fig.~\ref{fig_label}, there are imbalances in categories in the VEDAI dataset, and most of the targets are small targets, which further increase the challenge of detection.
Using the Mosaic data enhancement can increase the possibility of targets appearing in random positions and makes the target position distribution more uniform and dense, thereby solving the problem of uneven distribution. See the bottom left subgraph in Fig.~\ref{fig_label} for details.

\subsection{Ablation study}
Ablation experiments are performed on the VEDAI dataset for a detailed analysis in this section.
our baseline is the mono-modality detectors with RGB-only and IR-only, which is initialized from YOLOv5-small.

\subsubsection{Necessity of CMAFF module}

\begin{table*}[!htbp]
\caption{ Comparison of different models. Performance are evaluated in terms of AP and mAP.}\label{tab_performance}
\begin{tabular*}{\tblwidth}{@{}L@{}C@{}C@{}C@{}C@{}C@{}C@{}C@{}C@{}CL@{}}
\toprule
 Model & car & truck & pickup & tractor & camper & ship & plane & van & other & $\text{mAP}_{0.5}$ \\ 
 \midrule
 RGB-only           & 0.911 & 0.783 & 0.823 & 0.812 & 0.751 & 0.718 & 0.935 & 0.622 & 0.333 & 0.743 \\
 Thermal-only       & 0.867 & 0.771 & 0.759 & 0.623 & 0.666 & 0.707 & 0.995 & \textbf{0.843} & 0.430 & 0.740 \\
 Input Fusion I     & 0.887 & 0.777 & 0.795 & 0.723 & 0.757 &\textbf{0.786} & 0.957 & 0.721 & 0.295 & 0.744 \\
 Input Fusion II    & 0.891 & \textbf{0.849} & 0.787 & \textbf{0.842} & 0.701 & 0.699 & 0.984 & 0.688 & 0.389 & 0.760\\
 Mid Fusion I       & 0.882 & 0.849 & 0.817 & 0.684 & 0.722 & 0.710 & \textbf{0.995} & 0.536 & 0.536 & 0.748 \\
 Mid Fusion II      & 0.892 & 0.788 & \textbf{0.876} & 0.662 & 0.732 & 0.625 & \textbf{0.995} & 0.739 & 0.433 & 0.749 \\
YOLOFusion  & \textbf{0.917}& 0.781 & 0.859 & 0.719 & \textbf{0.789} & 0.711 & \textbf{0.995} & 0.752 & \textbf{0.547} & \textbf{0.786} \\
\bottomrule
\end{tabular*}
\end{table*}

\begin{figure*}[htbp]
    \centering
    \subfigure[Ground Truth]{
    \includegraphics[width=0.23\linewidth]{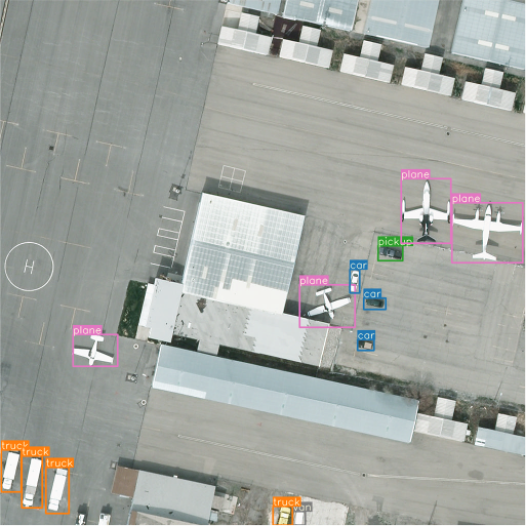}
    }
    \subfigure[Color-only]{
    \includegraphics[width=0.23\linewidth]{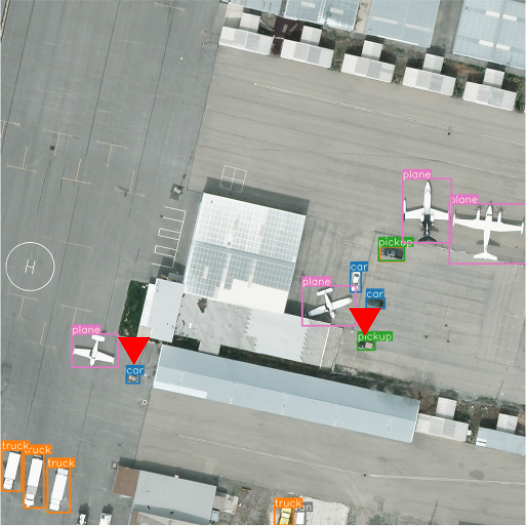}
    }
    \subfigure[Thermal-only]{
    \includegraphics[width=0.23\linewidth]{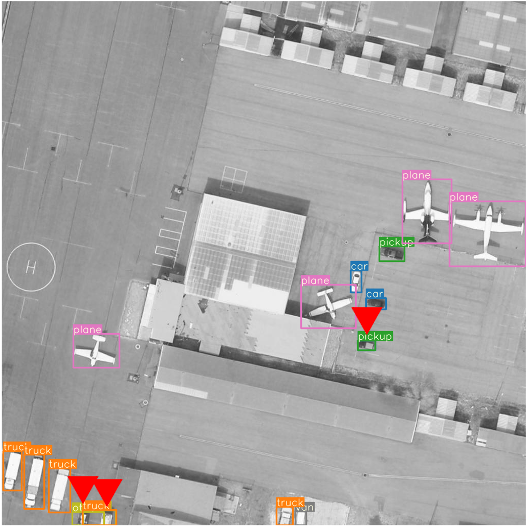}
    }
    \subfigure[Input Fusion I]{
    \includegraphics[width=0.23\linewidth]{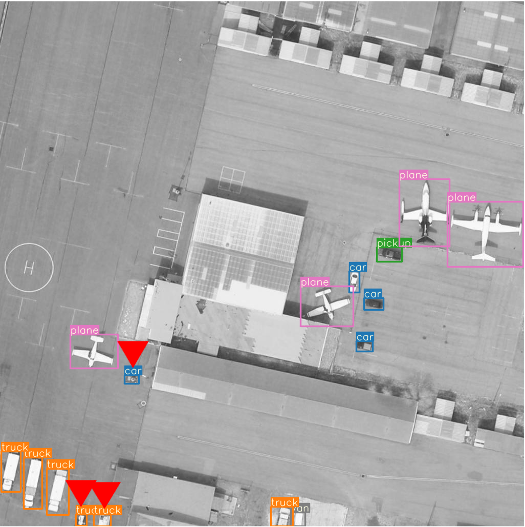}
    }
    
    \subfigure[Input Fusion II]{
    \includegraphics[width=0.23\linewidth]{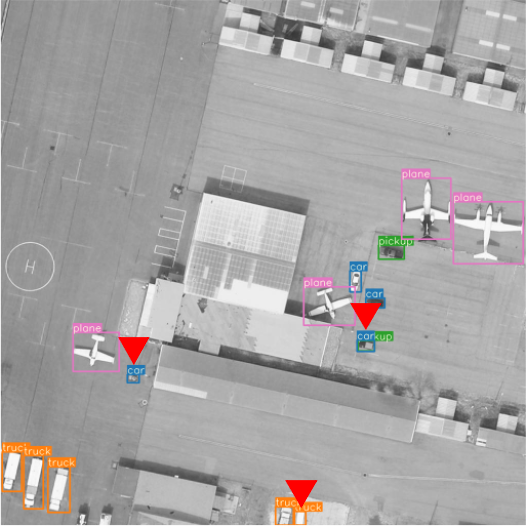}
    }
    \subfigure[Mid Fusion I]{
    \includegraphics[width=0.23\linewidth]{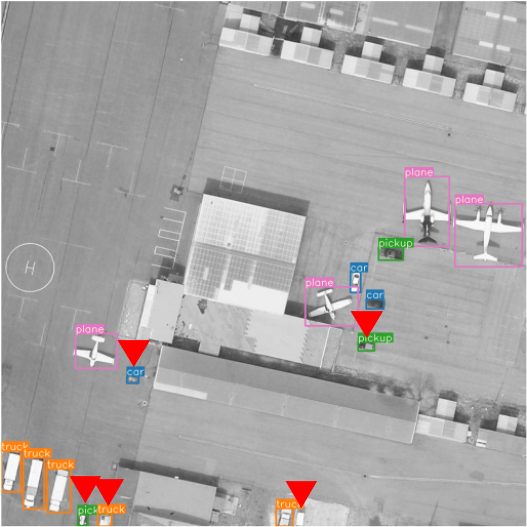}
    }
    \subfigure[Mid Fusion II]{
    \includegraphics[width=0.23\linewidth]{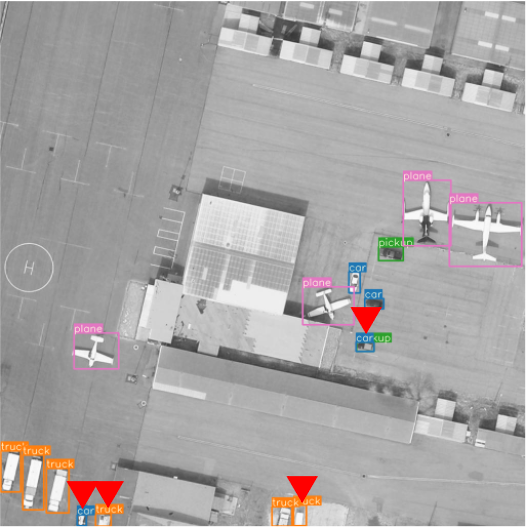}
    }
    \subfigure[Ours]{
    \includegraphics[width=0.23\linewidth]{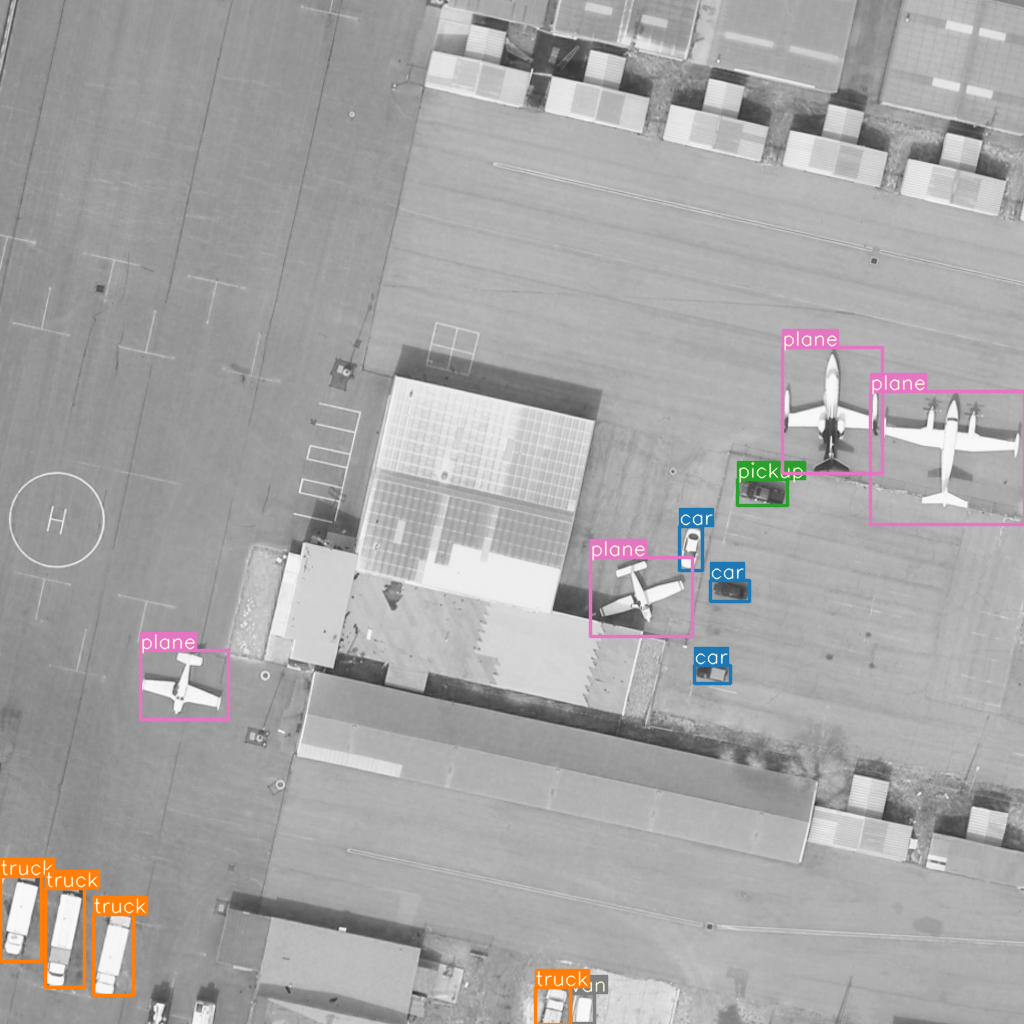}
    }

    \caption{
    Qualitative comparison of multispectral object detection in the VEDAI test subset with other approaches. (a) ground truth, (b) detection results of mono-modality detector trained by color dataset only, (c) detection results of mono-modality detector trained by thermal dataset only, (d) detection results of Input Fusion I, (e) detection results of Input Fusion II, (f) detection results of Mid Fusion I, (g) detection results of Mid Fusion II, (h) detection results of our method (YOLOFusion). 
    Note that {\color{red} red inverted triangles} indicate FPs and FNs.
    Zoom in to see details.
    }
    \label{fig_comparison}
\end{figure*}

Our fusion methods is compared with other approaches in Tab.~\ref{tab_performance}, and the mono-modality detectors (RGB-only and thermal-only) are implemented for our baselines.  
In addition, we provide two early fusion models (Input Fusion I and Input Fusion II) and two middle fusion models (Mid Fusion I and Mid Fusion II) for comparison. 
As mentioned in Fig.~\ref{fig_fusionform}, our Input Fusion I and Input Fusion II forms are pixel-level summation and channel-level concatenation, respectively. 
Likewise, Mid Fusion I and Mid Fusion II are both two-stream YOLOv5-small architectures as shown in Fig.~\ref{fig_yolofusion}, except for the CMAFF module which is replaced by summation and concatenation.

In Tab.~\ref{tab_performance},  the AP values of the nine classes are computed, and the best AP value of each item is in bold.
It can be seen that the mAP values of the RGB-only and thermal-only detection for the VEDAI dataset in the reasonable setting are 74.3\% and 74.0\%, both of which are very close.
Among our YOLOFusion and other four multi-modality fusion models, Input Fusion I with summation performed worst (74.4\%), while YOLOFusion performed best (78.6\%). 
The gap between YOLOFusion and others is at least 2.6\%, and up to 4.6\%. 
Specifically, YOLOFusion achieved the best APs, including car(0.917), camper(0.789) plane(0.995), and other(0.547).

We qualitatively compare our YOLOFusion with other methods in Fig.~\ref{fig_comparison}.
Visually, our YOLOFusion detects all objects, while the other methods have multiple false positives or false negatives , i.e., wrong detection.

\subsubsection{Impact of CMAFF architectures}\label{subsubsec_CMAFF_architectures}

Two experiment are discussed in this section, one is the impact of the differential enhancive module and the common selective module on CMAFF, and the other is how to arrange the two cross-modality modules to achieve the best performance.

\begin{table}[!htbp]
\caption{Comparing the impact of two cross-modality attention sub-modules on CMAFF. Performance are evaluated in terms of Precision, Recall and mAP.
DEM denotes the differential enhancive module and CSM denotes the common selective module.
}\label{tab_cmaff_archi}
\begin{tabular*}{0.5\linewidth}{@{}LC@{}C@{}C@{}C@{}L@{}}
\toprule
  &\multicolumn{2}{c}{CMAFF} & \multicolumn{3}{c}{Metric} \\ \cline{2-3} \cline{4-6} 
 Model &  DEM &   CSM &  Precision & Recall & $\text{mAP}_{0.5}$ \\ 
  \midrule
  Mid Fusion I &              &            & 0.716 &\textbf{0.725} & 0.748  \\
  Mid Fusion II &              &            & 0.749 & 0.717 & 0.749  \\
  YOLOFusion &\checkmark    &            & 0.778 & 0.690 & 0.757  \\
  YOLOFusion &              & \checkmark & 0.797 & 0.674 & 0.758  \\
  YOLOFusion &\checkmark    & \checkmark & \textbf{0.810} & 0.683 & \textbf{0.786}  \\
\bottomrule
\end{tabular*}
\end{table}
Table \ref{tab_cmaff_archi} shows the results of the effect of our two cross-modality modules. 
In this experiment, Mid Fusion I and II are the baseline for comparison.
According to the results, Using only the common selective module or the differential enhancive module can push performance near by 1\%.
Furthermore, the common-only manner performs slightly better than the differential-only manner.
And with the joint effect of the common selective module and the differential enhancive module, a more informative and robust fused feature map is generated for multispectral detection.

\begin{table}[!htbp]
\caption{Comparison of different CMAFF architectures. Performance are evaluated in terms of Precision, Recall and mAP.
DEM denotes the differential enhancive module and CSM denotes the common selective module.
}\label{tab_cmaff_archi2}
\begin{tabular*}{0.5\linewidth}{@{}LC@{}C@{}L@{}}
\toprule
    CMAFF &   Precision & Recall & $\text{mAP}_{0.5}$ \\ 
  \midrule
    CSM \& DEM in parallel            & \textbf{0.810} & 0.683 & \textbf{0.786}  \\
    CSM \& DEM(Concat) in parallel    & 0.760 & 0.675 & 0.768  \\
    CSM + DEM                         & 0.749 & \textbf{0.697} & 0.770  \\
    DEM + CSM                         & 0.791 & 0.690 & 0.763  \\
\bottomrule
\end{tabular*}
\end{table}

As each module has different functions, the order may affect the overall performance. 
In the second experiment, four different ways of arranging the common selective sub-module and the differential enhancive sub-module (sequential CSM-DEM, sequential DEM-CSM, parallel utilization of CSM and DEM, and parallel utilization of CSM and DEM(Concat)) are compared. 
Among them, the DEM(Concat) sub-module uses concatenation and 1x1 dimension-reduced convolution instead of the original final summation operation.

Table \ref{tab_cmaff_archi2} summarizes the experimental results on different attention arranging methods.
From the results, we can find that generating a parallel attention map infers a finer attention map than doing sequentially. 
In addition, the DEM performs better than the DEM(Concat) architecture, and CSM-first sequential order is superior to the DEM-first sequential order.
From the Tab.~\ref{tab_cmaff_archi} and Tab.~\ref{tab_cmaff_archi2}, it can be viewed that all the arranging methods outperform using only the common selective module or the differential enhancive module, showing that using both attentions is valuable while the best-arranging strategy further pushes performance.


\subsubsection{Attention mask interpretation}
\begin{figure*}[!htbp]
	\centering
		\includegraphics[width=\linewidth]{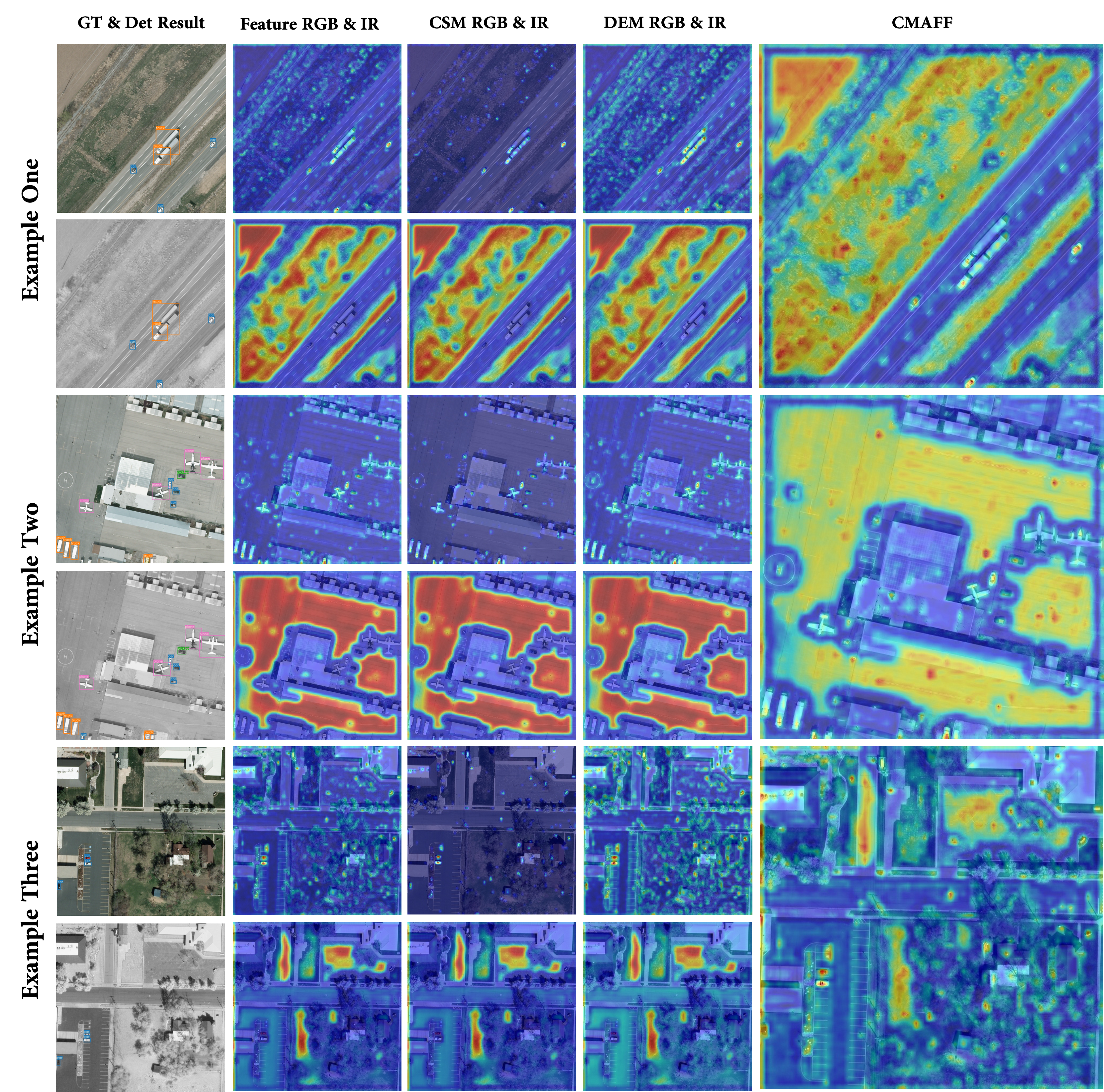}
	  \caption{Three examples of attention map visualization in CMAFF\_1 on the VEDAI dataset. 
	  The first column shows the ground truth and the detection result by YOLOFusion.
	  The second column shows feature maps of RGB and IR, i.e., the two-branch inputs of CMAFF\_1 modules.
	  The third column shows attention maps from the color branch and thermal branch of the common selective module.
	  The fourth column shows attention maps from the color branch and thermal branch of the differential enhancive module.
	  And the last column shows attention maps after the total CMAFF module.
	  Note that the top one means the color branch, and the bottom one represents the thermal branch in each example.
	  Zoom in to see details.
	  }
	  \label{fig_attention_mask}
\end{figure*}

Figure \ref{fig_attention_mask} provides the visualization results of the common selective, the differential enhancive, and the total CMAFF attention masks in three examples. 
Comparing the feature visualizations of the original RGB and common selective module in each example, it can be found that some redundant features are filtered by the common selective module, making the network focus on effective features.
Similarly, comparing the feature visualizations of the RGB and differential enhancive module, after passing through the differential enhancive module, the red area which focuses on the target becomes larger and the pixel color becomes much redder, that is, the feature is enhanced.
From these two comparisons, the functions of the two modules have been visually interpreted.

In addition, there is a very interesting point here that is the color-modality feature focuses on the small-area foreground, while the thermal-modality feature focuses on the large-area background. 
If the color-modality features infer a false positive that falls in the large-area foreground, utilizing the thermal-modality features by the CMAFF module, the false positive will be submerged in the foreground, thereby correcting this error.
This is equivalent to making a posterior correction for the color-modality features.

\begin{figure}[htbp]
    \centering
    \subfigure[Inputs features]{
    \label{subfig_input}
    \includegraphics[width=0.3\linewidth]{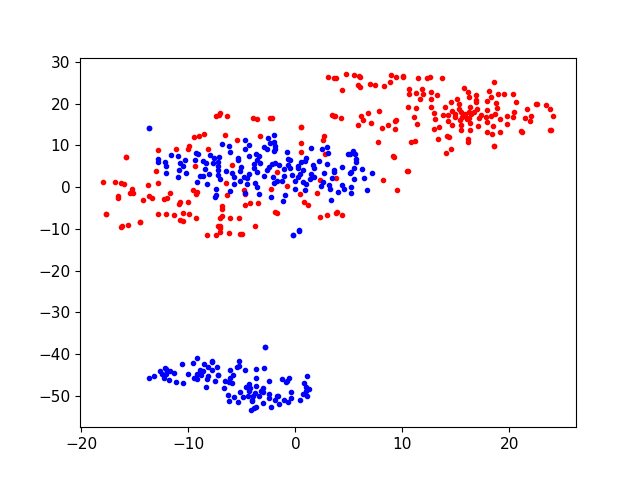}
    } 
    \subfigure[The common selective module]{
    \label{subfig_common}
    \includegraphics[width=0.3\linewidth]{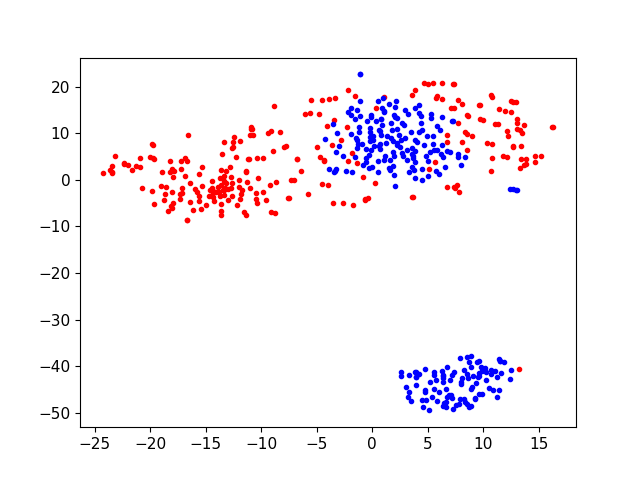}
    }
    \subfigure[The differential enhancive module]{
    \label{subfig_diff}
    \includegraphics[width=0.3\linewidth]{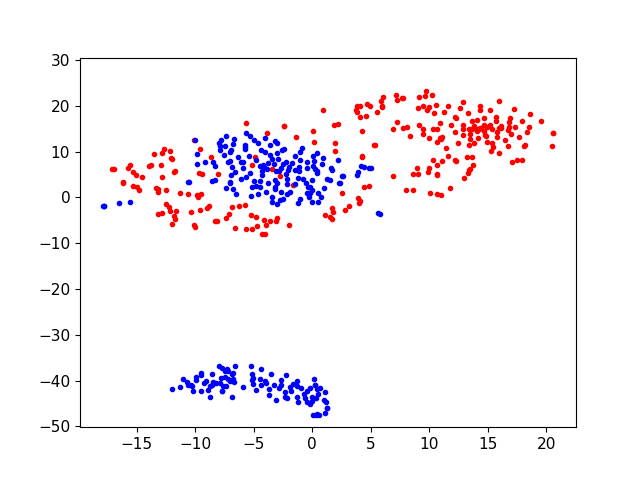}
    }
    \caption{Visualizing feature using t-SNE. The red dots represent the color-modality features, and the blue dots represent the thermal-modality features. }
    \label{fig_tsne}
\end{figure}

To further verify our idea, t-SNE~\cite{van2008visualizing} is used to visualize features including the input features of RGB and IR (the two-branch inputs of CMAFF\_2 module), the output features from the color branch and thermal branch of the common selective module, and the output features from the color branch and thermal branch of the differential enhancive module, in Fig.~\ref{fig_tsne}.
It can be seen that the thermal-modality features (blue dots) are naturally divided into two clusters.
We believe that the cluster overlapping with the red dots represents the foreground features learned by the thermal-branch network, while the cluster at the bottom, away from the red dots, is the background features. 
In other words, the color-modality features are all foreground features, and part of the thermal-modality features are foreground features and part of them are background features.
It is mutually confirmed with the view that the color-modality feature focuses on the small-area foreground, while the thermal-modality feature focuses on the large-area background in Fig.~\ref{fig_attention_mask}.
We guess that this phenomenon might be because the color-modality branch has a pre-trained model, which makes it have prior knowledge to learn the foreground features. 
In contrast, due to the lack of large-scale infrared data sets, the thermal-modality branch has no pre-trained models, or can only use the pre-trained color-modality models.
Therefore, in the multi-spectral learning process, the network is more inclined to let the color-modality branch learn foreground knowledge, and the thermal-modality branch tends to learn background knowledge.

Furthermore, the sub-figure~\ref{subfig_input} proves our initial assumption in section~\label{intro}, that is, the feature space of the two modalities has an overlapping modality-specific feature space.
It can be carefully observed from the sub-figure~\ref{subfig_common} that, compared to sub-figure~\ref{subfig_input}, near the cluster center of the blue dots in the overlapping area, the red dots are less distributed and tend to appear at the edge of the cluster. 
It indicates that our common selective module has a certain screening function for modality-shared features in the common feature space.
The feature distribution configuration of sub-figure~\ref{subfig_diff} and sub-figure~\ref{subfig_input} are roughly the same, but more closely.
After inferred by the differential enhancive module, the feature distribution has reasonable cohesion within classes (foreground and background) and separation between classes.

\subsection{Speed and parameter analysis}

\begin{table}[htbp]
\caption{Comparison of model parameters and runtime. 
The parameter unit M means million, and the runtime unit ms means millisecond.
The runtime x/x/x respectively represents the pure network inference time, the Non-Maximum Suppression (NMS) time and the total inference time.
}\label{tab_cmaff_params_runtime}
\begin{tabular*}{0.6\linewidth}{@{}L@{}C@{}C@{}CR@{}}
\toprule
 Model  & CMAFF & $\text{mAP}_{0.5}$ & Params(M) & Runtime(ms) \\ 
 \midrule
  Mid Fusion II &            & 0.749 & 11.97 & 4.3/1.5/5.8 \\
 YOLOFusion   & \checkmark & 0.786 & 12.52 & 4.4/1.5/5.9 \\          
 \midrule
 Margin     &            & 0.037 & 0.55 & 0.1/0/0.1 \\
\bottomrule
\end{tabular*}
\end{table}

In Tab.~\ref{tab_cmaff_params_runtime}, we report the total number of parameters and the average inference run-time.
Specifically, the models are implemented with Pytorch framework without TensorRT for an inference time testing on the NVIDIA$^\text{®}$ TITAN RTX™ platform. 
In the future, TensorRT can be used to further accelerate the models.
Our CMAFF involves only one sigmoid operation, one softmax layer and several convolutional layers with learn-able parameters, while the others are parameter-free, such as GAP, GMP, etc..
Since this, it increases only 0.55M of  parameters and 0.1ms of inference time.
Note that the additional parameters and inference time are caused by three CMAFF (CMAFF\_1,CMAFF\_2 and CMAFF\_3) modules.
Therefore the additional parameters of per module is only 0.183M on average.
It demonstrates that our CMAFF module is a lightweight cross-modality feature fusion design.

\subsection{Comparison with state-of-the-art object detection methods}
\tablename~\ref{tab_sota} shows the detection results of existing methods and our YOLOFusion on the VEDAI dataset. 
It can be observed that our YOLOFusion achieves state-of-the-art performance on this dataset. 
It is more accurate than the best mono-modality network YOLO-fine in $\text{mAP}_{0.5}$ ($2.86\%\uparrow$).
For multi-modality networks, the detection performance of our YOLOFusion greatly surpasses the previous network (YOLOv3 with mid-level fusion \cite{AAdhanaraj2020vehicle}) in a more stringent evaluation metric, $\text{mAP}_{0.5:0.95}$ ($4.52\%\uparrow$).
This is a very huge improvement, and it is worth noting that our YOLOFusion($\sim$12.5M) is a lightweight and small-scale network with far fewer parameters than the original YOLOv3($\sim$61.6M), let alone the dual-stream fused YOLOv3.

\begin{table}[htbp]
\caption{Comparison with the state-of-the-art object detection methods including mono-modality and multi-modality networks on VEDAI dataset.
}\label{tab_sota}
\begin{tabular*}{1.0\linewidth}{@{}L@{}C@{}C@{}C@{}C}
\toprule
 Model  & Dataset Type & Backbone &$\text{mAP}_{0.5}$ & $\text{mAP}_{0.5:0.95}$ \\ 
\midrule
 \multicolumn{5}{c}{mono-modality networks} \\
\midrule
Retina \cite{AAlin2017focal} & RGB & ResNet-50 & - & 0.4347 \\
Faster R-CNN \cite{AAren2015faster} & RGB & ResNet-101 & - & 0.3482 \\
SSSDET \cite{AAmandal2019sssdet} & RGB & shallow  network & - & 0.4597 \\
SSD \cite{AAliu2016ssd} & RGB & VGG16 & 0.7091 & - \\
SSD \cite{AAliu2016ssd} & IR & VGG16 & 0.6983 & - \\
EfficientDet(D0) \cite{AAtan2020efficientdet} & RGB &  EfficientNet B0 & 0.7068 & - \\
EfficientDet(D0) \cite{AAtan2020efficientdet} & IR &  EfficientNet B0 & 0.6979 & - \\
EfficientDet(D1) \cite{AAtan2020efficientdet} & RGB &  EfficientNet B1 & 0.7401 & - \\
EfficientDet(D1) \cite{AAtan2020efficientdet} & IR &  EfficientNet B1 & 0.7123 & - \\
YOLOv2 \cite{AAredmon2017yolo9000} & RGB & Darknet19 & 0.6572 & - \\
YOLOv2 \cite{AAredmon2017yolo9000} & IR & Darknet19 & 0.6316 & - \\
YOLOv3 \cite{AAredmon2018yolov3} & RGB & Darknet53 & 0.7311 & - \\
YOLOv3 \cite{AAredmon2018yolov3} & IR & Darknet53 & 0.7101 & - \\
YOLOv3-SPP \cite{AAredmon2018yolov3} & RGB & Darknet53 & 0.7504 & - \\
YOLOv3-SPP \cite{AAredmon2018yolov3} & IR & Darknet53 & 0.7370& - \\
YOLOv3-tiny \cite{AAredmon2018yolov3} & RGB & Darknet-tiny & 0.5555 & - \\
YOLOv3-tiny \cite{AAredmon2018yolov3} & IR & Darknet-tiny & 0.5536& - \\
YOLO-fine \cite{AApham2020yolo}  & RGB & lightweight Darknet53 & 0.7600 & - \\
YOLO-fine \cite{AApham2020yolo}  & IR & lightweight Darknet53 &   0.7517& - \\
YOLOv5-small \cite{AAjocher2020ultralytics} & RGB & CSPDarknet53 & 0.743 & 0.4624 \\
YOLOv5-small \cite{AAjocher2020ultralytics} & IR & CSPDarknet53 & 0.740 & 0.4621 \\
\midrule
 \multicolumn{5}{c}{multi-modality networks} \\
\midrule
 YOLOv3 with early fusion \cite{AAdhanaraj2020vehicle} & RGB+IR &  Darknet53 & - & 0.4403  \\ 
 YOLOv3 with mid-level fusion \cite{AAdhanaraj2020vehicle} & RGB+IR & two-stream Darknet53 & - & 0.4462  \\ 
 Input Fusion I (ours) & RGB+IR & CSPDarknet53 & 0.744 & 0.4563  \\  
 Input Fusion II (ours) & RGB+IR & CSPDarknet53 & 0.760 & 0.4807  \\  
 Mid Fusion I (ours) & RGB+IR & two-stream CSPDarknet53 & 0.748 & 0.4630  \\  
 Mid Fusion II (ours) & RGB+IR & two-stream CSPDarknet53 & 0.749 & 0.4658  \\  
 YOLOFusion (ours) & RGB+IR & two-stream CSPDarknet53 & \textbf{0.786} & \textbf{0.4914}  \\  
\bottomrule
\end{tabular*}
\end{table}

\section{Conclusion}\label{sec_conclu}
Following the idea of dealing with different features in different ways, we propose Cross-Modality Attentive Feature Fusion (CMAFF) module to fully exploit the inherent complementary between different modalities.
The common selective sub-module and the differential enhancive sub-module are two distinctive parts in the CMAFF to enhance and select the features, making detectors achieve considerable performance improvement while keeping the overhead small.
The different arrangements (parallel or sequential manner) of the two sub-modules are also explored to further boost the representation power of fused features.
We conduct extensive experiments on YOLOFusion to verify the efficacy of CMAFF.
Moreover, a visualization is provided to interpret how the module exactly fuses the given two branch features of different modalities.
Since CMAFF is a separate plug-in module, our module can be embedded in any type of two-stream CNN-based detectors as a fusion component with a little additional computation cost.

\section*{Acknowledgment}

This work was supported by the National Natural Science Foundation of China under Grant No.U20B2056 and No.11872034.











\bibliographystyle{cas-model2-names}

\bibliography{cas-refs}

\bio{}
\textbf{FANG Qingyun} was born in 1994. He received the bachelor’s degree in communication engineering from the Nanjing University of Science and Technology, in 2017. He is currently pursuing the Ph.D. degree with the School of Aerospace Engineering, Tsinghua University. His current research interests include deep learning and object detection.
\endbio

\bio{}
 \textbf{Wang Zhaokui} was born in Anhui, China, in 1978. He received the B.S. and Ph.D. degrees in space engineering from the National University of Defense Technology, Changsha, China, in 1999 and 2006, respectively. Since 2009, he has been working as an Associate Professor with Tsinghua University. He has authored or coauthored more than 100 journal articles and conference papers. He has authored one text book and coauthored two text books. He holds more than 20 patents in China and one patent in USA and one patent in Germany. His research aims at small satellite of scientific exploration and enabling technology of distributed space systems, such as dynamics and intelligent control for satellite cluster, inner-formation based gravity measurement satellite, human–robot collaborations, and multi-robots collaborations formations in space. He has been a Council Member of the Chinese Association of Automation (CAA) and a member of the American Institute of Aeronautics and Astronautics (AIAA) and the Space University Administrative Committee of International Astronautical Federation (IAF).
\endbio

\end{document}